\documentclass{article}
\usepackage{arxiv}
\usepackage{lmodern}
\usepackage[utf8]{inputenc} % allow utf-8 input
\usepackage[T1]{fontenc}    % use 8-bit T1 fonts
\usepackage{hyperref}       % hyperlinks
\usepackage{url}            % simple URL typesetting
\usepackage{booktabs}       % professional-quality tables
\usepackage{nicefrac}       % compact symbols for 1/2, etc.
\usepackage{microtype}      % microtypography
\usepackage{graphicx}
\usepackage[numbers, sort&compress]{natbib}
\usepackage{doi}
\usepackage[font=normalsize,labelfont={bf}]{caption}
\usepackage{subcaption}
\usepackage{graphicx}%
\usepackage{multirow}%
\usepackage{amsmath,amssymb,amsfonts}%
\usepackage{xcolor}%
\usepackage{textcomp}%
\usepackage{manyfoot}%
\usepackage{booktabs}%
\usepackage{csquotes}
\usepackage{mathtools}
\usepackage{siunitx}
\usepackage{tikz}
\usetikzlibrary{positioning}
\usetikzlibrary{shapes.geometric, arrows.meta}
\usepackage{definitions}

\usepackage[nameinlink]{cleveref}

\title{Physics-informed neural network estimation of active material properties in time-dependent cardiac biomechanical models}

\newif\ifuniqueAffiliation
\ifuniqueAffiliation % Standard variant of author block
\author{ \href{https://orcid.org/0000-0000-0000-0000}{\includegraphics[scale=0.06]{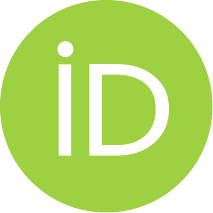}\hspace{1mm}David S.~Hippocampus}\thanks{Use footnote for providing further
		information about author (webpage, alternative
		address)---\emph{not} for acknowledging funding agencies.} \\
	Department of Computer Science\\
	Cranberry-Lemon University\\
	Pittsburgh, PA 15213 \\
	\texttt{hippo@cs.cranberry-lemon.edu} \\
	\And
	\href{https://orcid.org/0000-0000-0000-0000}{\includegraphics[scale=0.06]{orcid.pdf}\hspace{1mm}Elias D.~Striatum} \\
	Department of Electrical Engineering\\
	Mount-Sheikh University\\
	Santa Narimana, Levand \\
	\texttt{stariate@ee.mount-sheikh.edu} \\
}
\else
\usepackage{authblk}

\newbox{\orcid}\sbox{\orcid}{\includegraphics[scale=0.06]{orcid.pdf}} 
\author[1,3]{Matthias Höfler
}
\author[4]{%
	\href{https://orcid.org/0000-0002-4207-1400}{\usebox{\orcid}\hspace{1mm}Francesco Regazzoni}%
}
\author[4]{%
	\href{https://orcid.org/0000-0002-6662-3433}{\usebox{\orcid}\hspace{1mm}Stefano Pagani}%
}
\author[1,3]{%
	\href{https://orcid.org/0000-0002-4496-1933}{\usebox{\orcid}\hspace{1mm}Elias Karabelas}%
}
\author[2,3]{%
	\href{https://orcid.org/0000-0001-6341-4014}{\usebox{\orcid}\hspace{1mm}Christoph Augustin}%
}
\author[1,3]{%
	\href{https://orcid.org/0000-0002-3439-6117}{\usebox{\orcid}\hspace{1mm}Gundolf Haase}%
}
\author[2,3]{%
	\href{https://orcid.org/0000-0002-7380-6908}{\usebox{\orcid}\hspace{1mm}Gernot Plank}%
}
\author[1,2,3]{%
	\href{https://orcid.org/0000-0002-2637-0195}{\usebox{\orcid}\hspace{1mm}Federica Caforio\thanks{corr. author:~\texttt{federica.caforio@uni-graz.at}}}%
}
\affil[1]{Department of Mathematics and Scientific Computing, NAWI Graz, University of Graz, Austria}
\affil[2]{Gottfried Schatz Research Center: Division of Biophysics, Medical University of Graz, Austria}
\affil[3]{BioTechMed-Graz, Austria}
\affil[4]{MOX, Department of Mathematics, Politecnico di Milano, Milano, Italy}
\fi

\renewcommand{\shorttitle}{PINNs for active parameter estimation in biomechanical models}

\hypersetup{
pdftitle={PINNs for active parameter estimation in biomechanical models},
pdfauthor={Federica Caforio et al},
pdfkeywords={First keyword, Second keyword, More},
}

\begin{document}
\maketitle

\begin{abstract}
	Active stress models in cardiac biomechanics account for the mechanical deformation caused by muscle activity, thus providing a link between the electrophysiological and mechanical properties of the tissue. The accurate assessment of active stress parameters is fundamental for a precise understanding of myocardial function but remains difficult to achieve in a clinical setting, especially when only displacement and strain data from medical imaging modalities are available. 
This work investigates, through an in-silico study, the application of physics-informed neural networks (PINNs) for inferring active contractility parameters in time-dependent cardiac biomechanical models from these types of imaging data. 
In particular, by parametrising the sought state and parameter field with two neural networks, respectively, and formulating an energy minimisation problem to search for the optimal network parameters, we are able to reconstruct in various settings active stress fields in the presence of noise and with a high spatial resolution. 
To this end, we also advance the vanilla PINN learning algorithm with the use of adaptive weighting schemes, ad-hoc regularisation strategies, Fourier features, and suitable network architectures.
In addition, we thoroughly analyse the influence of the loss weights in the reconstruction of active stress parameters.
Finally, we apply the method to the characterisation of tissue inhomogeneities and detection of fibrotic scars in myocardial tissue.
This approach opens a new pathway to significantly improve the diagnosis, treatment planning, and management of heart conditions associated with cardiac fibrosis.
\end{abstract}

% keywords can be removed

\section{Introduction}
Developments in precision medicine increasingly rely on the integration of patient-specific data and computational models to improve the diagnosis and treatment of cardiovascular pathologies. 
Despite technological progress, optimising cardiovascular therapies remains an unsolved challenge, primarily due to the complex nature of myocardial function.
%Continuing from our investigation into the passive properties of the myocardium, this study focuses on the estimation of active myocardial properties. 

A critical aspect of cardiac function is the active contractility of the myocardium, i.e., the heart's ability to generate force and contract. 
In pathological conditions such as myocardial infarction, regions of the heart lose this contractile ability, forming scars that significantly impair cardiac function.
The accurate identification and characterization of these regions is essential for diagnosis, treatment planning, and prognosis. 
However, directly measuring active contractility properties is not feasible in clinical settings, creating a critical need for computational methods that can reliably estimate these properties from available imaging data, such as echocardiography, computed tomography, or MRI images, solving ad hoc inverse problems. 
%In this context, a challenging task is the estimation of active contractility properties of cardiac tissue.
%These properties cannot be directly measured and therefore must be inferred from imaging data such as echocardiography, computed tomography, or MRI images, solving ad hoc inverse problems.
The challenge of quantifying active contractility has already been explored in the literature with data assimilation and partial differential equations (PDEs)-constrained optimisation strategies.
In~\citep{sermesant2006cardiac}, the authors used a data assimilation framework to estimate ventricular contractility from MRI images.
Optimisation-based approaches were analysed by~\citep{sun2009computationally}, targeting global minimum convergence issues in the estimation of regional contractility properties from 3D-tagged MRI.
Further efforts in this domain include the works of~\citep{finsberg2018estimating}, in which estimation methods were derived based on echocardiographic data, and~\citep{asner2016estimation}, in which 3D-tagged magnetic resonance MRI data was used to provide information on myocardial motion. 
%The foundational model of Guccione et al. for material property estimation has underpinned much of the subsequent research in this area, followed by the contributions from Augenstein et al. and Wang et al.
Recent contributions to extract local contractility properties based on subdivision in AHA regions using data assimilation strategies include~\citep{chabiniok2012estimation} and~\citep{imperiale2021sequential}, considering clinical data from cine-MRI and tagged-MRI, respectively.
Despite demonstrated accuracy, these methods often rely on significant computational resources and prior assumptions, and they typically condense the complex mechanics into a reduced set of parameters. 
Recently,~\citep{kovacheva_estimating_2021} provided a methodology to estimate the active tension field in the myocardium considering in silico wall motion and Tikhonov regularisation.
Another approach to estimate the active contractility field based on PDE-constrained optimisation is given in~\citep{pozzi2024reconstruction}, using in silico displacement data in a quasi-static framework.
To the best of our knowledge, no specific method relying on machine learning has been developed to estimate active properties in cardiac tissue.
In this work, we address this challenge through a novel approach, based on physics-informed neural networks (PINNs),  that enables the estimation of spatially-varying active contractility properties both in quasi-static and time-dependent scenarios from limited displacement and strain measurements, potentially enabling more accurate cardiac assessment in clinical practice.
%In this work, we employ Physics-Informed Neural Networks (PINNs) to efficiently and accurately estimate heterogeneous active contractility properties.  
PINNs utilise a combination of data-driven and physics-based principles, which can circumvent the high computational cost of conventional optimisation or inverse problems methods. 
The integration of this approach with cardiac biomechanical models aims at providing a more precise estimation of heterogeneous active myocardial properties and thus improving the personalisation of such models.
Building upon our previous work on estimating passive properties in cardiac biomechanical models~\citep{caforio2024physics}, we extend the original approach introduced in~\citep{Raissi2019} to enable distributed parameter estimation in soft tissue nonlinear biomechanics, particularly when tissue properties are not known \emph{a priori} and may vary spatially.
In this work, we advance beyond passive mechanics to address the more complex challenge of active stress parameter estimation, which introduces fundamental time-dependencies and physiological complexities. 
This necessitates extending our analysis to time-dependent biomechanical models of cardiac function.
Moreover, we consider scenarios characterised by limited data that reflect real-world clinical applications where only displacement and strain information are available, f.e., extracted from cine-MRI or tagged-MRI images, with no corresponding stress data. 
Our approach is based on the use of ad-hoc optimisation techniques, regularisation strategies and network architectures and allows for the estimation of spatially-varying parameters using only a small set of (sparse) displacement and, in some cases, strain data, and it requires fewer epochs and training data compared to standard PINN approaches for inverse problems in elasticity~\citep{Haghighat2021a,kamali2023elasticity}. 
Our method is particularly effective in estimating heterogeneous tissue properties without necessitating prior assumptions.
We accurately reconstruct scars' shapes without relying on stress data, also when the data has a low resolution and is only accessible on parallel slices (such as MRI image stacks). 
To the best of our knowledge, no other method offering similar performance has been documented in the existing literature.
The proposed results are consistent with detailed finite element simulations across various test cases, encompassing both healthy and pathological scenarios. 
Moreover, we demonstrate the robustness of our model predictions in terms of parameter estimation, reconstruction of displacement, and detection of scarred tissue, against noise. 
Additionally, we thoroughly analyse the influence of regularisation, hard constraints on the boundaries, and hyperparameter choice on the precision and accuracy of predictions.
Regarding the analysis of hyperparameters' impact on PINN prediction, we conduct a thorough investigation of the apparent Pareto front~\citep{rohrhofer_data_2023} between data fidelity and PDE residual losses in the context of scalar parameter estimation, identifying optimal hyperparameter settings that balance the trade-off between physical consistency and adherence to observational data and provide the most accurate estimation of the sought parameters.
The rest of the manuscript is structured as follows: \Cref{sec:methods} is devoted to the description of the inverse problem strategy based on PINNs for estimating constant and space-dependent active stress parameters in soft tissue nonlinear biomechanical models considering passive and active stress, including detailed information on the optimisation approach, the regularisation strategies, hyperparameter tuning based on the analysis of the apparent Pareto fronts and the network architecture. 
\Cref{sec:Results} includes numerous test cases where we show the performance of the methodology with different degree of complexity in terms of heterogeneity of the parameters to be estimated, and we study the influence of noise on the accuracy of the inferred results.
In \Cref{sec:Discussion}, we examine the characteristics and potential future directions of the methodology, while \Cref{sec:Conclusion} presents concluding remarks on this study.

\section{Methods}
\label{sec:methods}
In this section, we describe our comprehensive approach for estimating active cardiac tissue properties using PINNs. 
We first present the general PINN framework for parameter estimation, introducing the cardiac biomechanical model governing tissue behaviour. 
Next, we detail our optimisation strategy and specialised regularisation techniques designed to address the specific challenges of active property estimation. 
Finally, we describe our network architecture and enhancements concerning hyperparameter tuning that improve performance for heterogeneous tissue characterisation. 
Each component is designed to contribute to the accurate reconstruction of spatially-varying contractility parameters from limited measurements.

\subsection{Parameter Estimation with PINNs}
In this work, we adopt the PINN framework to estimate active stress parameters in three-dimensional, time-dependent  cardiac biomechanical models. 
\subsubsection{Governing equations in cardiac biomechanics}
\begin{figure}[ht]
    \centering
   \includegraphics[width=0.5\textwidth]{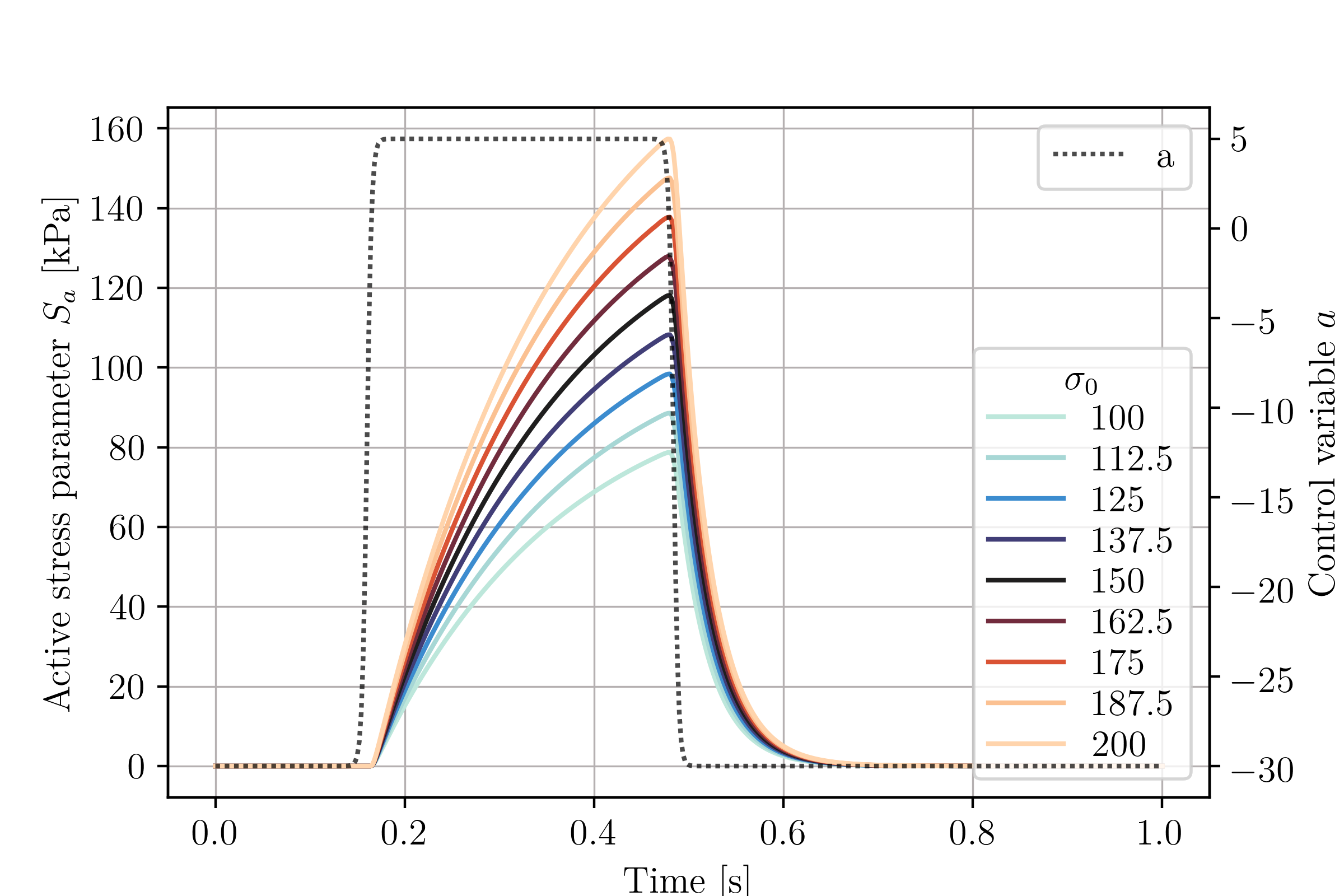}
    \caption{Time-evolution of the Bestel-Clément-Sorine model with different values of the maximal active stiffness $\sigma_0>0$.}
    \label{fig:bestel_ode}
\end{figure}

The starting point is Cauchy's equation in the time-dependent formulation,
\begin{equation}
    \begin{cases}
	\rho\partial_t^2 \bu - \nabla \cdot \bP(\bu) = \bZero, & \text{in} \ \Omega \\
	\bP(\bu) \, \vn = - p \, J \, \bF^{-\top}\, \vn & \text{on} \ \Gamma_N\\ 
	\bu = \bg & \text{on} \ \Gamma_D \\
    \bu(0) = \bu_0, \, \partial_t \bu(0) = \bv_0 & \text{in} \ \Omega,
    \end{cases}
\label{eq:pde}
\end{equation}
where $\bu(\bX, t) \in \R^3$ describes the displacement at the point $\bX \in \R^3$ and time $t>0$ in Lagrangian formulation, $\bF=\bI + \nabla \bu$ is the deformation tensor, $J = \det(F)$ is its Jacobian, and $\bP(\bu)$ the first Piola-Kirchhoff stress tensor.
In this work we consider zero body forces.
$\Gamma_N$ represents the portion of the boundary where Neumann boundary conditions (BCs) are imposed, whereas Dirichlet BCs are imposed on $\Gamma_D$.  
The unit vector $\bn$ denotes the outward normal vector on $\Gamma_N$ and $p$ is a given pressure.
The vectors $\bu_0$ and $\bv_0$ represent the initial condition for the displacement and velocity, respectively.
We consider an active stress formulation of $\bP(\bu)$ in the form
\[
\bP(\bu) = \bP_{pas}(\bu) + \bP_{act}(\bu),
\]
where $\bP_{pas}(\bu)$ models passive behaviour and $\bP_{act}(\bu)$ active stress triggered by electrophysiological activation via excitation-contraction coupling mechanisms.
According to standard assumptions in soft tissue modelling, particularly in the cardiac setting~\citep{Holzapfel2009Constitutive}, the passive tissue behaviour is modelled as a nearly incompressible~\citep{Flory1961}, hyperelastic material. 
Its mechanical response is derived from an associated strain energy function $\mathcal{W}=\mathcal{W}(\bF)$:
\[
\bP = \frac{\partial \mathcal{W}}{\partial \bF}.
\]
We consider the transverse-isotropic Guccione material model~\citep{Guccione1991Passive}
\[
\mathcal{W}=\frac{\alpha_P}{2}(\exp (\overline{Q})-1)+\frac{\kappa}{2}(\log J)^2,
\]
where
\[
\begin{aligned}
\overline{Q} = & b_{\mathrm{f}}\left(\mathbf{f}_0 \cdot \overline{\mathbf{E}} \mathbf{f}_0\right)^2+ \\
& b_{\mathrm{t}}\left[\left(\mathbf{s}_0 \cdot \overline{\mathbf{E}} \mathbf{s}_0\right)^2+\left(\mathbf{n}_0 \cdot \overline{\mathbf{E}} \mathbf{n}_0\right)^2+2\left(\mathbf{s}_0 \cdot \overline{\mathbf{E}} \mathbf{n}_0\right)^2\right]+\\
& 2 b_{\mathrm{fs}}\left[\left(\mathbf{f}_0 \cdot \overline{\mathbf{E}} \mathbf{s}_0\right)^2+\left(\mathbf{f}_0 \cdot \overline{\mathbf{E}} \mathbf{n}_0\right)^2\right],
\end{aligned},
\]
with $\bmf_0$ the myocyte fibre orientation; $\bs_0$ the sheet orientation; $\bn_0$ the sheet-normal orientation.
Moreover, $\overline{\mathbf{E}}=\frac{1}{2} \left( \overline{\bC} - \bI\right)$ denotes
the isochoric Green–Lagrange strain tensor, where $\overline{\mathbf{C}}:=J^{-2 / 3} \bF^T\bF$ is the isochoric right Cauchy-Green deformation tensor. 
The parameter $\alpha_P$ denotes the passive stiffness and is set to $\SI{0.8}{\kPa}$, whereas the bulk modulus $\kappa$ is set to $\SI{650}{\kPa}$.
Default values of $b_\mathrm{f} = 18.48$, $b_\mathrm{t} = 3.58$, and $b_\mathrm{fs} = 1.627$ are used.
For the active stress part, we consider the model~\citep{ambrosi2012active}
\[
\mathbf{P}_{\mathrm{act}}=S_a(t) \frac{\mathbf{F} \mathbf{f}_0 \otimes \mathbf{f}_0}{\sqrt{\mathbf{F} \mathbf{f}_0 \cdot \mathbf{F} \mathbf{f}_0}},
\]
which acts along the fibre direction $\bmf_0$.
In this work we consider a combination of Dirichlet (BCD) and Neumann (BCN)  boundary conditions, as detailed in~\Cref{sec:Results}.\\ 
The time-dependent active stress parameter $S_a(t)$ translates the electrophysiological signal into mechanical stress. It is modelled by the Bestel-Clément-Sorine model~\citep{bestel2001biomechanical}
\begin{equation} \label{eqn:bestel-clement-sorine}
\left\{\begin{array}{l}
\dot{S}_a(t)=-|a(t)| S_a(t)+\sigma_0|a(t)|_{+} \, \text { for } t \text { in }(0, T] \\
S_a(0)=0,
\end{array}\right.
\end{equation}
where the additional control variable $a(t)$ is given by~\citep{AROSTICA2025117485}
\[
\begin{aligned}
|a(t)|_{+} & =\max \{a(t), 0\} \\
a(t) & =\alpha_{\max } f(t)+\alpha_{\min } \cdot(1-f(t)) \\
f(t) & \left.=S^{+}\left(t-t_{\text {sys }}\right) \cdot S^{-}\left(t-t_{\text {dias }}\right)\right) \\
S^{ \pm}(\Delta t) & =\frac{1}{2}\left( \pm \tanh \left(\frac{\Delta t}{\gamma}\right)\right).
\end{aligned}
\]
The parameters $\alpha_{max}, \alpha_{min}, t_{sys}, t_{dias}$ are related to the cardiac cycle. Physiological values $\alpha_{min}=-30$, $\alpha_{max}=5$, $t_{sys}=\SI{0.161}{\s}$, $t_{dias}=\SI{0.484}{\s} $~\citep{AROSTICA2025117485} are used.
From a biomechanical point of view, the maximum active stiffness $\sigma_0 \geq 0$ gives most information about the biomechanical properties of the tissue.
Also, if $S_a^1(t)$ denotes the solution to the initial-value problem \eqref{eqn:bestel-clement-sorine} with $\sigma_0=1$, then the rescaled version $S_a^\sigma(t) \coloneqq \sigma S_a^1(t)$ solves the same initial-value problem \eqref{eqn:bestel-clement-sorine} with parameter $\sigma_0=\sigma$, as depicted in~\Cref{fig:bestel_ode}. 
In this work, we consider both a quasi-static approach and the time-dependent problem.
In the first setting, for a fixed time point $t^* \in (0,T]$, we denote the corresponding displacement $u^*(\bX) \in \R^3$ as the solution to the quasi-static equation
\[
- \nabla \cdot \bP_*(\bu^*) = \bmf,
\]
where
\[
\bP_*(\bu) = \bP_{pas}(\bu) + S_a(t^*) \frac{\mathbf{F} \mathbf{f}_0 \otimes \mathbf{f}_0}{\sqrt{\mathbf{F} \mathbf{f}_0 \cdot \mathbf{F} \mathbf{f}_0}}.
\]
Provided that we can compute $S_a^1(t^*)$, if we know $S_a(t^*)$, we can immediately reconstruct 
\[
\sigma_0 = \frac{S_a(t^*)}{S_a^1(t^*)}.
\]
For cardiac tissue, $\sigma_0 >0$ is expected to vary in space depending on the health state of the tissue. This makes the solution of system \eqref{eqn:bestel-clement-sorine} for the time-trajectory of $S_a(t)$ also space-dependent. Therefore, in general, we write $\sigma_0(\bx)$ and $S_a(t; \bx)$.
%
%In the following sections we will consider several test cases with boundary conditions
%
\subsubsection{Solving inverse problems in cardiac biomechanics with PINNs}
\label{sec:methods_IP}
The PINN framework enables to approximate a solution to a given PDE and identify unknown parameters encoding the fundamental physical law governing a phenomenon partially measured with sensors. 
In our in-silico test cases, the observation data employed to train the neural network is the FEM numerical solution of the test case considered,
%with spatial resolution equal to \siunit{0.2\mm}, 
uniformly sampled in the domain to obtain $\bu_i^{\text{obs}}$, for $i = 1,2,\ldots, N_\mathrm{obs}$. 
We mimic the presence of measurement error by corrupting this data with additive white noise, i.e. zero-mean Gaussian noise with a given standard deviation $\sigma$:
\begin{equation}
	\label{eq:noisy_meas}
	\tilde{\bu} = \bu + \bvareps , \quad \bvareps \sim \cN(0,\sigma^2).
\end{equation}
As in~\citep{caforio2024physics}, we consider a metric defined as limiting dispersion (LD), defined as:
\begin{equation*}
	\text{LD} = \frac{3\sigma}{\max({\bu})}.
\end{equation*}
Specifically, training the network involves minimising a carefully designed cost function that considers the residuals of the governing PDE, the initial and boundary conditions, and the data discrepancy term. 
We refer to~\citep{caforio2024physics} for further detail on the basic principles of PINNs and our implementation for applications in soft tissue mechanics.
%\subsubsection{PINNs - a short review}
\paragraph{Homogeneous case}
First, we consider the case of a homogeneous parameter field $\sigma_0(\bx)=\sigma_0$ in the domain, i.e., a constant parameter. 
For the quasi-static test case, this means that we want to reconstruct $S_a(t^*; \bx) = S_a(t^*) \eqqcolon S_a$.
We represent the displacement field through a NN taking as input the three spatial coordinates and outputs the three components of the displacement field on that point. 
The resulting problem is the following optimisation problem:
\emph{Find the weights and biases $\hat\bw$ of an artificial neural network $\mathrm{NN}_{\bu}$ and the unknown parameter $\hat S_a$  s.t. :}
\begin{equation}
    \label{eq:pinn}
    \begin{split}
	\hat\bw, \hat S_a &= \underset{\bw,S_a}{\operatorname{argmin}} \big(\bcJ_{\text{OBS}}(\bw) +\bcJ_{\text{PDE}}(\bw; S_a) \\ & \quad  + \bcJ_{\text{BC}}(\bw) + \bcR(\bw)\big),
    \end{split}
\end{equation}
\emph{where the mean squared error loss functions $\bcJ_{*}$ and regularisation term $\bcR(\bw)$ read:}
\begin{equation}
   \label{eq:loss}
\begin{aligned}
	\hspace{-2ex}\bcJ_{\text{OBS}}( \bw) &=  \frac{\lambda_{\text{OBS}}}{N_\mathrm{obs}}\sum_{i = 1}^{N_{\text{obs}} }\norm{\bu_i^{\text{obs}} - \mathrm{NN}_{\bu}(\bx_i^{\text{obs}}; \bw)}^2,\\
	\hspace{-2ex}\bcJ_{\text{PDE}}(  \bw; S_a) &= \\ &\hspace{-6.5ex}\frac{\lambda_{\text{PDE}}}{N_\mathrm{pde}}\sum_{i = 1}^{N_{\text{pde}} }\norm{\bbf(\mathbf{x}^{\text{pde}}_{i}) - \bcL\bigl(\mathrm{NN}_{\bu}(\mathbf{x}^{\text{pde}}_i;  \bw);S_a\bigr)}^2,\\
    \hspace{-2ex}\bcJ_{\text{BC}}(\bw) &= \bcJ_{\text{BCD}}(\bw) + \bcJ_{\text{BCN}}(\bw), \\
	\hspace{-2ex}\bcJ_{\text{BCD}}(\bw) &= \\
    &\hspace{-6.5ex} \frac{\lambda_{\text{BCD}}}{N_\mathrm{bc}}\sum_{i = 1}^{N_{\text{bcd}} }\norm{\bu_{\Gamma_D} (\mathbf{x}^{\text{bcd}}_{i}) - \mathrm{NN}_{\bu}(\mathbf{x}^{\text{bcd}}_i; \bw)}^2,\\
    \hspace{-2ex}\bcJ_{\text{BCN}}(\bw) &= \\
    &\hspace{-6.5ex} \frac{\lambda_{\text{BCN}}}{N_\mathrm{bcn}}\sum_{i = 1}^{N_{\text{bcn}} }\norm{
    \mathbf p(\mathbf{x}^{\text{bcn}}_i) - \bP\left(\mathrm{NN}_{\bu}(\mathbf{x}^{\text{bcn}}_i; \bw)\right)\mathbf n}^2, \\
 \hspace{-3ex}\bcR(\bw) &= \lambda_{w} \norm{\bw}^2,
\end{aligned}
\end{equation}
\emph{with
%$\bu_i^{\text{obs}}$ , for $ i = 1, ..., N_{\text{obs}}$, represent the available observations on displacement data, whereas 
$\{\bx_i^\text{pde}\}_{i=1}^{N_\text{pde}}$, $\{\bx_i^\text{bcd}\}_{i=1}^{N_\text{bcd}}$, and $\{\bx_i^\text{bcn}\}_{i=1}^{N_\text{bcn}}$ denoting the collocation points for the PDE residual loss and the BCD and BCN loss terms, respectively.}
The Neumann data $\mathbf p$ is a pre-described pressure on the respective boundaries.
To ensure clarity, we avoid using subscript 2 when referring to the $l^2$-norm.
Hyperparameters $\lambda_i$, for $i\in\{\text{OBS},\text{PDE},\text{BCD},\text{BCN}, w$\}, play a role in non-dimensionalising each loss term and in weighting the contribution of each term to the overall loss function.
%The inputs to $\mathrm{NN}_{\bu}$ consist of point coordinates, while the outputs yield displacement vectors computed at those input locations.
Given that the parameter $S_a$ can also be defined as trainable parameters, the framework inherently enables us to conduct parameter estimation (model inversion) \citep{Raissi2019,Haghighat2021a}.
\paragraph{Heterogeneous case}

A more general approach consists in treating the parameter $S_a$ as a field $S_a(\mathbf{x})$, i.e. $S_a \colon \Omega \to \R$. 
To do so, instead of considering the parameter $S_a$ as a trainable variable (constant in space), we simultaneously train two neural networks, $\mathrm{NN}_{\bu}\big(\bx;\bw_1\big)$ for displacement and $\mathrm{NN}_{S_a}(\bx; \bw_2)$ for the parameter, to solve the following minimisation problem: \\
\emph{Find the weights and biases $\hat\bw_1$, $\hat\bw_2$ of two artificial neural networks $\mathrm{NN}_{\bu}, \ \mathrm{NN}_{S_a}$ s.t.:}
\begin{equation}
\label{eq:PINN_field}
\begin{split}
\hspace{-2mm}
   \hspace{-2ex}\hat\bw_1, \hat\bw_2 =&  \underset{ \bw_1,  \bw_2}{\operatorname{argmin}} \big(\bcJ_{\text{OBS}}( \bw_1)+ \bcJ_{\text{PDE}}(\bw_1;  \bw_2)   \\ &\hspace{-2ex}  +  \bcJ_{\text{BC}}(\bw_1;  \bw_2) + \bcR_1(\bw_1) + \bcR_2(\bw_2) \big),
    \end{split} 
\end{equation}
\emph{where we have introduced additional regularisation terms for the two networks $\mathrm{NN}_{\bu}, \ \mathrm{NN}_{S_a}$.}
We remark that, with an abuse of notation, we employ the same notation $\bcJ_{\text{PDE}}$ and $\bcJ_{\text{BC}}$ to denote the physics-informed loss functions in both cases, although its definition differs between the homogeneous and the heterogeneous case.
\subsection{Optimisation scheme}
The optimisation procedure involved in Eqs.~\eqref{eq:pinn} and \eqref{eq:PINN_field} is done considering the following two-step approach.
First, we perform a pre-training using only the reduced loss term $\bcJ_{\text{OBS}}( \bw_1)$.
We employ 600 iterations of the ADAM optimiser, followed by a BFGS optimisation phase until convergence to a local minimum. 
Then, we perform a full training, based on Eqs.~\eqref{eq:pinn} and \eqref{eq:PINN_field}, respectively.
This training involves an initial phase with $N_\mathrm{ADAM}$ iterations using the ADAM optimiser, followed by subsequent $N_\mathrm{BFGS}$ iterations of BFGS optimisation (the exact values of $N_\mathrm{ADAM}$ and $N_\mathrm{BFGS}$ depend on the complexity of the test case considered) an will be later specified.
\subsection{Regularisation}
Regularisation is applied for both the displacement network $\mathrm{NN}_{\bu}$ and the active stress network $\mathrm{NN}_{S_a}$, depending on the test case.

\subsubsection{Weight decay}
\label{weight_decay}
Weight decay is a standard regularisation strategy for neural networks~\cite[Chapter 7]{Goodfellow-et-al-2016}. 
It consists of penalising the $l^2$-norm of the weights $\bw$ of the neural network, i.e.:
\begin{equation}
     \bcR_1(\bw) = \lambda_{w} \norm{\bw}^2,
\end{equation}
with a suitably chosen hyperparameter $\lambda_{w}$ tuning the influence of the regularisation.
However, in the context of PINNs, their use has been discouraged already for forward problems~\citep{wang_experts_2023}.
In this work we have experimented weight decay for both networks. 
However, it did not show a beneficial effect on the reconstruction of either the displacement or the active stress field, but instead introduced an additional hyperparameter.  
%\textcolor{Matthias}{add small comparison in the appendix, this would be helpful to demonstrate this sentence.}
Hence, we incorporate it only for the initial, simpler test cases in~\Cref{sec:Results}, and there only for the displacement network.
\subsubsection{Regularisation for the active stress network}
\label{sec:reg}
As stated in Appendix~\ref{sec:appendix_identifiability}, the identification of active stress parameters near the boundary might be problematic. 
In particular, for the case of homogeneous Dirichlet boundary conditions
\[
\bu = 0 \quad \text{ on } \Gamma_D \subset \partial \Omega,
\]
not only the displacement vanishes but also the gradient of the displacement network $\text{NN}_{\bu}(\cdot; \bf{w}_1)$ tends to become small, and also the gradient of $\text{NN}_{S_a}(\cdot; \bf{w}_2)$, as detailed in \cref{eqn:Saexplicit} in Appendix \ref{sec:appendix_identifiability}. 
To mitigate this problem, we introduce a weight function $\omega: \partial \Omega \longrightarrow [0,1]$ with the properties that $\omega(\bx) \approx 0$ near the Dirichlet boundary $\Gamma_D$ and $\omega(\bx)$ approaches $1$ elsewhere. 
The modified boundary loss term reads then
\[\begin{aligned}
\bcJ_{\text{BCN}}(\bw) &= \\
&\hspace{-8.5ex}
\frac{\lambda_{\text{BCN}}}{N_\mathrm{bcn}}\!\sum_{i = 1}^{N_{\text{bcn}} }
\!\omega(\bx_i^{\text{bcn}})
\norm{
    \mathbf p(\mathbf{x}^{\text{bcn}}_i) - \bP\!\left(\mathrm{NN}_{\bu}(\mathbf{x}^{\text{bcn}}_i; \bw)\right)\! \mathbf n}^2,
\end{aligned}
\]
which effectively reduces its influence near the Dirichlet boundary~\citep{pozzi2024reconstruction}. 
In order to compensate for this loss of information, we add additional regularisation on the parameter network $\text{NN}_{S_a}(\cdot; \bf{w}_1)$. 
Let us denote with $\bar \omega = 1 - \omega$ the complementary weight. 
Then, the effect of the additional regularisation should be scaled according to $\bar \omega$. 
As a baseline, we choose a gradient based penalty that was already used in~\citep{pozzi2024reconstruction}, modified by the complementary weight:
\begin{equation*}
\begin{split}
&\hspace{-3mm}\bcR_2(\bw_2) = \\
&\quad \frac{1}{N_\mathrm{bc}} \sum_{i=1}^{N_{\text{bc}} } \bar \omega(\bx_i^{\text{bc}}) \norm{\nabla \text{NN}_{S_a}(\mathbf{x}^{\text{bc}}_i; \bf{w}_1) }^2.
\end{split}
\end{equation*}
In Appendix~\ref{sec:appendix_reg} we show the results of the estimation with and without this regularisation on the active stress network.
\subsection{Hyperparameter tuning}
One of the most challenging parts of the successful implementation of PINNs is the choice of the loss weights $\lambda_i$, $i \in \{ \text{OBS}, \text{PDE}, \text{BC}, \text{w}\}$. 
%As can be seen in Figure \textcolor{Matthias}{reference figure}, t
The presence of noise makes the choice of weights even more crucial. It influences both the forward and the inverse problem as illustrated in~\Cref{fig:pareto}.
In Appendix~\ref{sec:appendix_adaptive_weights} we make a thorough analysis between a fixed choice of static weights and more advanced adaptive weighting schemes.
\subsubsection{Analysis of the apparent Pareto fronts}
\label{sec:pareto_front}
Viewing PINNs from a multi-objective optimisation point of view, the aim is to minimise several targets, namely the fulfilment of the PDE, the boundary conditions, and the accordance to the given data. 
In most cases, all these targets are represented by a discrepancy and are combined with a linear weighting scheme. 
The overall PINN performance depends crucially on this scalarisation of the multi-objective problem. 
In \citep{rohrhofer_data_2023}, it was shown that not only the choice of the weights for each target, but also the parameters of the physical system under study have an influence on the performance of the method. 
This study is based on the analysis of the apparent Pareto front of the physical system, which can be defined as the set of loss values achievable with gradient-based training and enables to visualise the effect of scaling on multi-objective optimisation.
In particular, loss weights can counterbalance scaling effects of the system parameters.
In addition, certain system parametrisations leading to more balanced residuals can introduce locally convex regions, enabling successful gradient-based training across a broader spectrum of loss weights.
It remains unclear how the choice of weights influences inverse problems within the framework of PINNs.
In order to shed light on this aspect, 
%in \Cref{sec:res_static} 
we analyse the apparent Pareto fronts for a representative test case in the context of parameter estimation with PINNs. 
Our analysis provides practical insights on PINNs revealing a range of optimal hyperparameter values that yield the best parameter estimation performance and illustrating the trade-off between data fidelity and PDE adherence in cardiac tissue models.
\subsubsection{Adaptive weighting}
\label{sec:adaptive_weights}
Inspired by arguments given in \citep{wang_experts_2023}, we investigate the use of loss balancing terms. 
In that work, the authors compared two adaptive weighting schemes: a simple approach based on the gradients of the loss functions and a more sophisticated one based on the neural tangent kernel. 
They observed similar overall performance of both methods. However, the former showed more fluctuations, potentially caused by mini batching, while the latter had higher computational complexity.
Since we are using full batches also during the Adam phase, we can prevent these effects, hence we choose the more lightweight gradient based scheme. 
Let us denote $i,j \in \{ \text{OBS}, \text{PDE}, \text{BC}\}$, and the weight and loss values at iteration $k \in \N$ with $\lambda_i^k$ and $\bcJ_i^k(\bw)$ respectively. We first compute the global weights $\hat\lambda_i$ via the balance property
\[
\hat\lambda_i = \frac{1}{\norm{\nabla_{\bw}\bcJ_i^k(\bw)}} \sum_j \norm{\nabla_{\bw}\bcJ_j^k(\bw)}, \quad  \forall  i.
\]
Then, we compute the update $\lambda_i^{k+1}$ using a moving average of the form
\[
\lambda_i^{k+1} = \alpha \hat\lambda_i + (1-\alpha) \lambda_i^k,
\]
where $0<\alpha<1$ is a predefined hyperparameter controlling the moving average.

\subsubsection{Residual-Based Attention mechanism}
\label{sec:RBA}
We also incorporate a Residual-Based Attention (RBA) mechanism as introduced by \citep{anagnostopoulos_residual-based_2024}. 
The basic idea is to introduce a space-dependent weighting scheme for a particular loss.
Let us denote with $e^k(\bx)$ the error for a particular objective function at a given point $\bx$ at iteration $k \in \N$ in the optimisation process. 
For example, for the PDE it takes the form \[e^k(\bx)=\norm{\bbf\left(\mathbf{x}\right) - \bcL\left(\mathrm{NN}_{\bu}\left(\mathbf{x};  \bw^k\right);S_a\right)}^2,\] 
where $\bw^k$ denotes the parameter vector of the neural network $\mathrm{NN}_{\bu}$ after iteration $k$. 
We define the weight of the PDE loss for the next iteration $k+1$ as
\[
\lambda^{k+1}(\bx) = \gamma \lambda^k(\bx) + \eta^* \frac{e^k(\bx)}{\norm{e^k}_\infty} + \lambda_0,
\]
where the decay parameter $\gamma>0$, the learning rate $\eta^*>0$, and the lower bound $\lambda_0 \geq 0$ are given hyperparameters. 
For the initialisation, we choose $\lambda^0(\bx)=0$. 
The modified loss now reads
\begin{align*}
\bcJ_{\text{PDE}}(  \bw; S_a) = \frac{\lambda_{\text{PDE}}}{N}\sum_{i = 1}^{N} \lambda(\bx_{i}) e(\bx_{i}),
\end{align*}
where we have omitted the dependence on the iteration step for brevity.
%
%\textcolor{Matthias}{Changed: updated testcases for homogeneous problem also incorporate RBA.}
Instead of updating the weights on every iteration, we choose an update interval of five iterations for the sake of efficiency.
Comparison of results using no weight balancing, adaptive weighting, or RBA is shown in Appendix~\ref{sec:appendix_adaptive_weights}.
%We use RBA only for the test cases where also the parameter is considered as a distributed field given by a neural network as a compromise between enhanced accuracy and higher computational complexity.
%
\subsection{Network architecture}
The baseline architecture for the displacement network $\text{NN}_{\bu}(\cdot; \bf{w}_1)$ and the active stress network 
%(when the active stress parameter is considered as a heterogeneous field)
$\text{NN}_{S_a}(\cdot; \bf{w}_2)$ consists of two fully connected feed-forward neural networks. 
Both networks use $\tanh$ as activation function.
The number of layers and neurons depends on the test case considered and is specified in~\Cref{sec:Results}.
We consider a standard Xavier uniform initialisation of the network weights and biases.
To ensure robust predictions for both displacement reconstruction and parameter estimation, we evaluate multiple random initialisations of neural network weights and biases using different seeds and present the geometric mean and trajectory-spanned area in the results.
\subsubsection{Interval constraints}
In order to ensure that the PDE formulation is well-posed and to incorporate additional knowledge explicitly into the network structure, we consider ad-hoc activation functions for the last layer for the active stress network $\text{NN}_{S_a}(\cdot; \bf{w}_2)$. 
In the homogeneous case we define this function as the quadratic function $f(x)=x^2$, to ensure non-negativity.
In the heterogeneous case, we set interval constraints in the form
\[
f(x) = \left( S_{a,\max} - S_{a, \min} \right) \sigma(\alpha_S x) + S_{a, \min},
\]
where $S_{a,\max}>0$ is related to the active contractility of healthy tissue, $S_{a, \min}>0$ is a small value, and $\sigma(x)$ denotes the sigmoid function where the additional scaling parameter $\alpha_S >0$ modifies the slope of the sigmoid function.
\subsubsection{Exact incorporation of boundary conditions}
\label{sec:exact_BCD}
In order to alleviate the limitation of multi-objective minimisation problems, some boundary conditions can also be imposed explicitly instead of considering residual terms in the loss function.
We follow the approach proposed in~\citep{sukumar_exact_2022} to explicitly encode boundary conditions in neural networks. 
Especially for the case of a Dirichlet boundary described by
\[
\bu (\bx) = \bm{g} (\bx) \quad \text{ on } \Gamma_D \subset \partial \Omega,
\]
we modify the network structure as follows. 
Let $\phi_{\Gamma_D}: \R^3  \longrightarrow \R$ be a (signed) distance function such that the zero-level-set contains only the Dirichlet boundary portion, i.e. 
$$\Gamma_D = \{ \bx \in \bar\Omega \mid \phi_{\Gamma_D}(\bx)=0 \}.$$
% $\Gamma_D \subset \{ \bx \in \R^3 \mid \phi_{\Gamma_D}(\bx)=0 \}$ and $\Bar{\Omega} \setminus \Gamma_D \not \subset \{ \bx \in \Omega \mid \phi_{\Gamma_D}(\bx)=0 \}$. 
We further denote by $\text{NN}(\cdot; \bf{w}): \R^3 \longrightarrow \R^3$ an arbitrary neural network. 
The modified neural network with the explicitly incorporated Dirichlet boundary is then defined via
\[
\widetilde{\text{NN}}(\bx; \bf{w}) \coloneqq \phi_{\Gamma_D}(\bx) \cdot \text{NN}(\bx; \bf{w}) + \bf{g}(\bx),
\]
where $\bf{g}$ has been suitably extended to the whole domain.
Appendix~\ref{sec:appendix_reg} shows the results of the estimation using weakly enforced Dirichlet BC, whereas Appendix~\ref{sec:appendix_BC_robin} deals with the robustness of the PINN prediction in the presence of misspecification of boundary conditions.
\subsubsection{Fourier features}
Several works, e.g. \citep{ramasinghe2022frequency, cao_towards_2020, tancik2020fourier, wang_when_2022}, reported issues of neural networks learning high-frequency features. 
\citep{xu_frequency_2020} showed empirically that neural networks tend to learn the lower parts of the frequency spectrum of the target function faster. 
In \citep{wang_when_2022}, the neural tangent kernel theory was utilised to explain this so-called \textit{spectral bias}, i.e.,  the tendency of neural networks to learn low-frequency components faster than high-frequency ones. 
This spectral bias poses a particular challenge for the parameter field network in active property estimation, as myocardial scars and tissue heterogeneities manifest as high-frequency spatial components that require precise characterisation, although the associated displacement field is typically smooth.
Therefore we use Fourier feature embeddings for the parameter field network only. 
In more detail, the Fourier feature mapping can be viewed as an additional layer transforming the spatial input coordinates into frequency space. 
For a given a-priori chosen frequency $\sigma_F>0$ and feature dimension $m \in \N$, we sample elements $b_{ij} \sim \mathcal{N}(0, \sigma_F)$ and compose them in a non-trainable feature matrix $\bm{B} \in \R^{m \times 3}$. 
The physical input variable $\bx$ is then transformed in the frequency space according to
\[
\gamma(\mathbf{x})=\left[\begin{array}{l}
\cos (\mathbf{B x}) \\
\sin (\mathbf{B x})
\end{array}\right].
\]
In \citep{wang_experts_2023}, a discussion on the effect and training sensitivity of the hyper-parameter $\sigma_F$ is given. 
In particular, the authors observed that a too low value can lead to even more emphasis on low-frequency solutions, leading to blurry results, whereas high values of $\sigma_F$ can lead to salt-and-pepper artifacts. 
The need for a careful tuning of the Fourier feature frequency was also emphasised in \citep{tancik2020fourier}. 
Also in this work, we observe salt-and-pepper artifacts when using a too high frequency, leading to a relatively small choice for $\sigma_F \in [1,3]$.

\section{Results}
%%%
% to generate schematic NN : http://alexlenail.me/NN-SVG/index.html
\label{sec:Results}
%
%In what follows, we consider different test cases concerning different patterns for the constant or spatially-dependent active stress parameter $S_a(\bx)$. 
In what follows, we present our results in order of increasing complexity to systematically demonstrate the capabilities of our approach. We begin with homogeneous test cases using quasi-static approximations, where active stress is constant throughout the domain. 
We then progress to time-dependent models that capture the cardiac dynamics. 
Finally, we address the most challenging scenario of heterogeneous active stress fields, including the detection of scars with different geometries and configurations. 
This progression allows us to validate each aspect of our methodology and demonstrates its applicability to increasingly realistic cardiac modelling scenarios.
The considered geometry is described by a cube of side length $\SI{10}{\mm}$ centred in [0,0,0]\,\SI{}{\mm}, corresponding to a small specimen of cardiac tissue.
The unit vectors of the fibre, sheet and normal directions are, respectively, $[1,0,0]$, $[0,1,0]$ and $[0,0, 1]$.
We also enforce homogeneous Dirichlet boundary conditions on the face $y=\SI{-5}{\mm}$, and homogeneous Neumann boundary conditions on all the other faces of the computational domain, respectively. 
The BCD are exactly incorporated as detailed in \Cref{sec:exact_BCD}, whereas the BCN are imposed in terms of the residual loss expressed in Eq.~\eqref{eq:loss}.
%Finally, we assume that there are no body forces, i.e., $\bmf = \bZero$.\\
The training dataset used in the observation loss of Eq. \eqref{eq:loss} is represented by \emph{in silico} data randomly sampled from the solution of the high-fidelity FEM simulator \href{https://carpentry.medunigraz.at/index.html}{\texttt{carpentry}}~\cite{Augustin2016,caforio2021coupling,Karabelas2022}.
The open-source software \href{https://carpentry.medunigraz.at/getting-started/carputils-overview.html}{\texttt{carputils}} is used to define input/output tasks and feature definition and extraction, e.g. definition of tagged regions on meshes with different parameters.
In all the test cases presented, we consider RBA for the PDE loss, as described in \Cref{sec:RBA}. This choice is justified in Appendix~\ref{sec:appendix_adaptive_weights}.
\subsection{Homogeneous test case}
In this section, we evaluate the performance of the method in reconstructing a displacement field and estimating a scalar contractility parameter using only random point-wise measurements of the displacement field corrupted by different noise levels. 
\subsubsection{Quasi-static approximation}
\label{sec:res_static}
\begin{figure}[h!]
    \includegraphics[width=\linewidth]{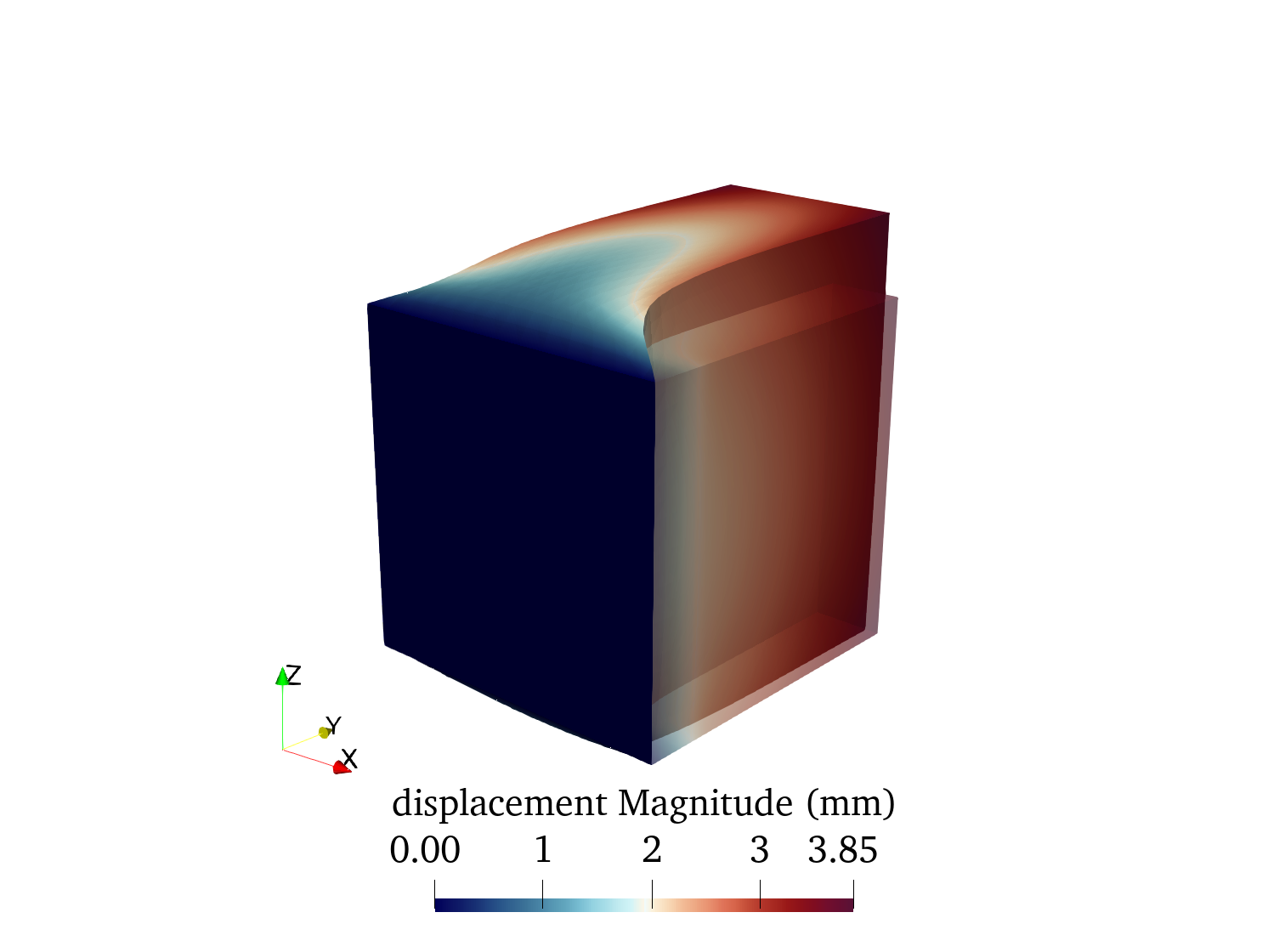}
    \caption{Ground-truth FEM solution of the displacement field. The light-shaded cube represents the reference configuration, whereas the coloured object is in deformed configuration. %Colour coding is according to the displacement magnitude.
    }
    \label{fig:grid}
\end{figure}

\begin{figure*}[h]
    \centering
    % First subfigure
    \begin{subfigure}[b]{0.49\textwidth}
        \centering
        \includegraphics[width=\textwidth]{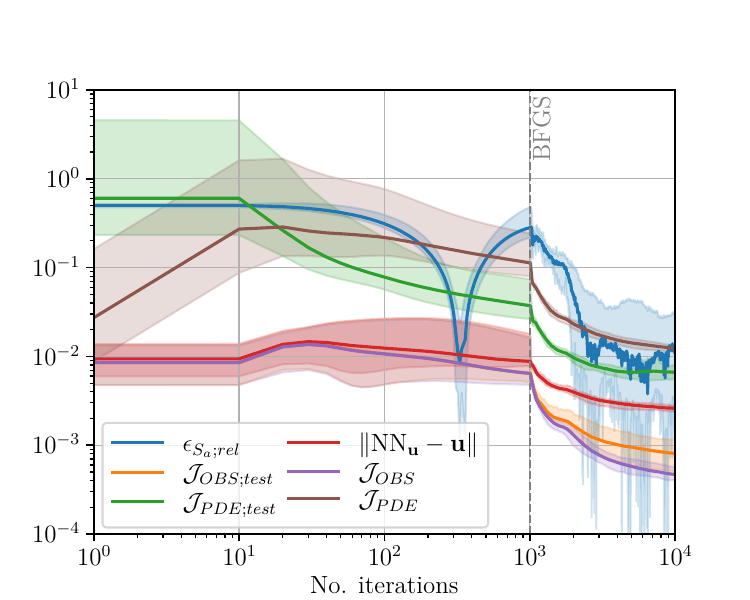}
        \caption{$LD=0$.}
    \end{subfigure}
    % Second subfigure
    \begin{subfigure}[b]{0.49\textwidth}
        \centering
        \includegraphics[width=\textwidth]{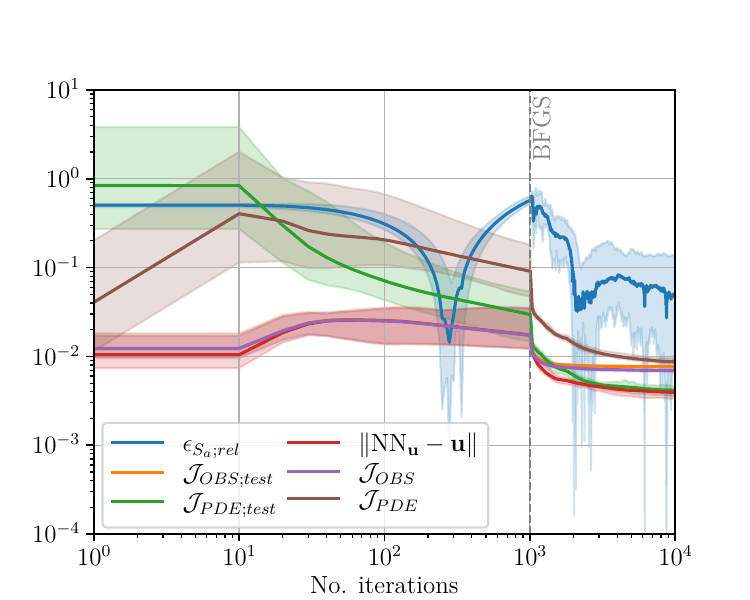}
        \caption{$LD=0.05$.}
        \label{fig:T1_hom_single:subnoise}
    \end{subfigure}
    \caption{Comparison of training and testing losses for displacement data and PDE discrepancy and the relative error on the parameter $S_a$. Left: Algorithm performance considering noiseless data. Right: Performance using data corrupted with noise corresponding to $LD = 0.05$. The solid line depicts the geometric mean over the seeds; the shaded region is the area spanned by the trajectories. The second dashed vertical line marks the selected end of training at 10k BFGS epochs, for which we report performance and associated errors. For completeness, we also display the algorithm's behavior up to 50k epochs.}
    \label{fig:T1_hom_single}
\end{figure*}

\begin{figure*}[h]
    \centering
    \begin{subfigure}{\linewidth}  % Width of each subfigure
        \centering
        \includegraphics[width=\linewidth]{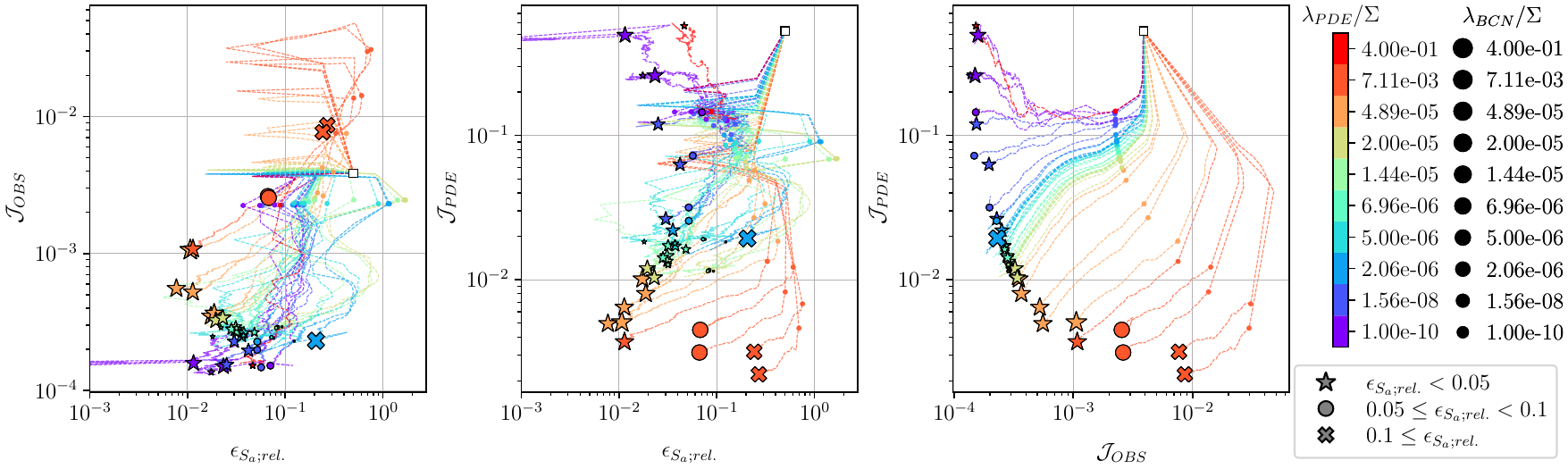}
        \caption{$LD=0.00$}  % Subfigure caption
        \label{fig:pareto:000}
    \end{subfigure}
    \hfill
    \begin{subfigure}{\linewidth}  % Width of each subfigure
        \centering
        \includegraphics[width=\linewidth]{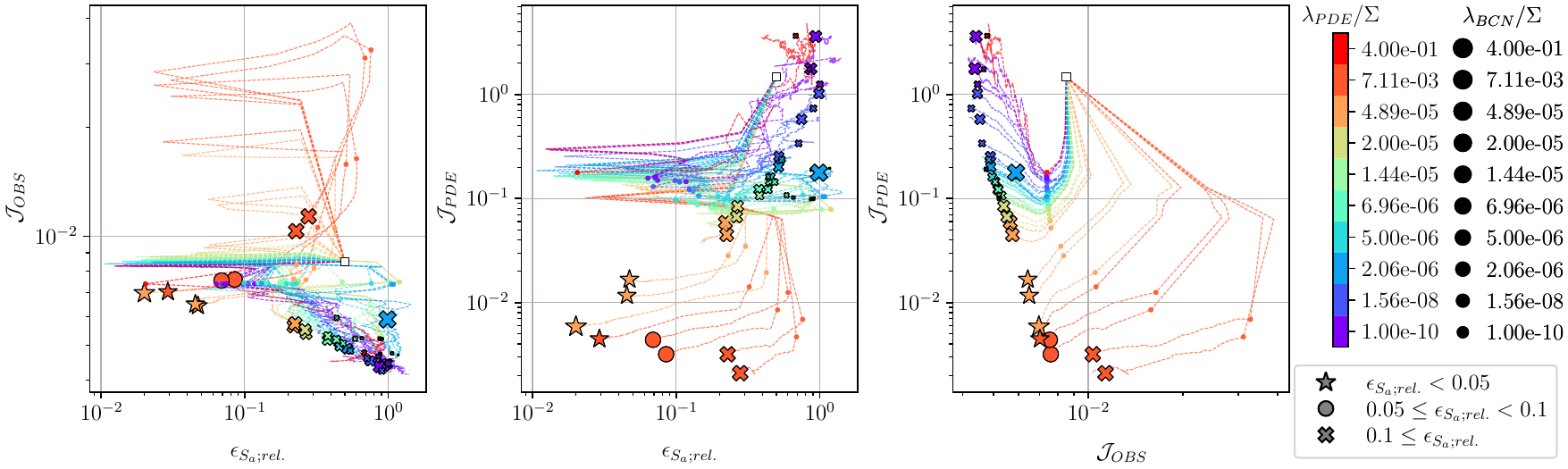}
        \caption{$LD=0.05$}  % Subfigure caption
        \label{fig:pareto:005}
    \end{subfigure}
    \caption{Analysis of apparent Pareto fronts (constant active stress parameter $S_a$, quasi-static approximation). 
    %Pareto front for the single parameter case with $LD=0.05$. 
    The results show the different training trajectories $\bcJ_{\text{PDE}}$ and $\bcJ_{\text{BCN}}$ and the relative error in the parameter $S_a$, denoted by $\epsilon_{S_a; rel}$, with different weights for $\bcJ_{\text{OBS}}$, $\bcJ_{\text{PDE}}$, and $\bcJ_{\text{BCN}}$. For each weight combination, the sum $\Sigma=\lambda_{\text{OBS}}+\lambda_{\text{PDE}}+\lambda_{\text{BC}}$ is computed, and, based on this normalising factor, the fraction of each of the three weights $\lambda_i$, $i \in \{ \text{OBS}, \text{PDE}, \text{BC}\}$, is then given by $\lambda_i/\Sigma$. The colour encodes the fraction of the chosen PDE weight whereas the size of the end marker encodes the fraction of the BCN weight. Every trajectory represents an average over three different seeds. All trajectories start at the same configuration (indicated by a white square). The small coloured circle along the trajectory indicates the transition from Adam to BFGS optimisation. The endpoints are marked by different symbols according to the endvalue of the relative error $\epsilon_{S_a; rel}$: a star if $\epsilon_{S_a; rel} < 0.05$, a circle if $0.05 \leq \epsilon_{S_a; rel} < 0.1$, and a cross if $\epsilon_{S_a; rel}>0.1$. The noiseless case is shown in \Cref{fig:pareto:000} whereas in \Cref{fig:pareto:005} we consider $LD=0.05$ as a noise level.
    }.
    \label{fig:pareto}
\end{figure*}

%Here, we consider the case of a constant parameter $S_a$ that is modelled with a single trainable parameter. 
%Here, we consider the case of a constant parameter $S_a$. 
We consider the case when we know a priori that  $S_a$ is constant, and therefore we model it with a single trainable parameter.
The ground-truth solution for $S_a=\SI{118}{\kilo\pascal}$ is given in~\Cref{fig:grid}.
The training is composed by \num{1000} Adam epochs followed by \num{10000} BFGS epochs. 
%However, the BFGS iterations stopped earlier for every instance leading to an average amount of \num{39560} and \num{38780} BFGS epochs for the cases $LD=\{0.00, 0.05\}$, respectively. 
%In the noiseless case, two out of ten seeds failed, i.e., they stopped right after the Adam iterations.
\Cref{tab:T1_hom_single} depicts the relative $L^2$ testing error for the state $\bu$ and relative error on $S_a$ w.r.t to the ground-truth value, whereas~\Cref{fig:T1_hom_single} depicts the training and testing losses for the displacement data and the PDE residual, together with the relative error on $S_a$, showing low generalisation error and the robustness of the method predictions also in presence of noise.
\begin{table}[ht]
\centering
\begin{tabular}{cccc} 
\toprule
\textbf{$LD$} & \textbf{no. seeds} & \textbf{$L^2$ rel. err. on $\bu$} & \textbf{$\epsilon_{S_a; rel}$}\\
\midrule
0.00 & 10 & \num{3.21e-02} & \num{1.24e-02} \\ \midrule
0.05 & 10 & \num{4.84e-02} & \num{4.93e-02} \\
\bottomrule
\end{tabular}
\caption{Performance for the quasi-static case, with $S_a$ modelled as a constant parameter. 
The first column indicates the noise level, the second column the number of successful seeds. The test case is repeated with ten seeds in total. 
The third column shows $\| \text{NN}_{\bu} - \bu \|/\|\bu \|$, the $L^2$ relative error for the state $\bu$ evaluated on the ground-truth mesh. 
The last column shows the relative error on $S_a$ w.r.t the ground-truth value. 
Both errors are first averaged over the different seeds and then averaged over the values from \num{9900} to \num{10000} iterations.}
\label{tab:T1_hom_single}
\end{table}
\paragraph{Analysis of apparent Pareto front}
% Mephisto:
%Pareto_Front_Ambrosi_SNR5_BL_samestart/Data/numPDE2500numBCN100numData500adam1600bfgs1200adam3600bfgs360000wPDE2.0e+00wFit2.0e+01wBCN1.0e+00wT1.0e-08seed1BLSNR5.0e-02alpha0_0_0_.json
We perform the analysis of the Pareto fronts to study the influence of weight selection on the training of the PINN and the relative error on the prediction of $S_a$.
\Cref{fig:pareto} shows the results of this analysis considering noiseless data and data affected by noise corresponding to $LD = 0.05$. 
Both in the noiseless and in the noisy cases it is possible to observe a convex front in the rightmost plot, comparing the training trajectories of $\bcJ_{\text{PDE}}$ and $\bcJ_{\text{OBS}}$, with final values of the PDE training loss negatively correlated to the relative weight given to the respective loss, as one could expect.
The sweet spot according to the Pareto front is reached with the following relative weights (normalised by the sum) given in~\Cref{tab:pareto_front}.
\begin{table*}[h]
\centering
\begin{tabular}{ccccc} 
\toprule
\textbf{$LD$} & \textbf{$\lambda_{\text{OBS}}/\Sigma$} & \textbf{$\lambda_{\text{PDE}}/\Sigma$} & \textbf{$\lambda_{\text{BCN}}/\Sigma$} & \textbf{$\epsilon_{S_a; rel}$} \\ \midrule
$0.00$ & \num{9.985e-01} & \num{9.885e-04} & \num{4.993e-04} & \num{7.71e-03} \\ \midrule
$0.05$ & \num{9.852e-01} & \num{4.926e-03} & \num{9.842e-03} & \num{2.00e-02} \\
%$0.05$ & \num{0.9852314} & \num{0.004926157} & \num{0.009842462} \\
\bottomrule
\end{tabular}
\caption{Analysis of apparent Pareto fronts (constant active stress parameter $S_a$, quasi-static approximation). Optimal weights obtained from the Pareto front analysis in~\Cref{fig:pareto}. The normalising factor is given by $\Sigma = \lambda_{\text{OBS}} + \lambda_{\text{PDE}} + \lambda_{\text{BCN}}$.}
\label{tab:pareto_front}
\end{table*}
\paragraph{Learning a spatially-constant active stress network}
Here we consider again the case of constant $S_a$. 
However, unlike in Sec.~\ref{sec:res_static}, we assume that its homogeneity in space is unknown.
%, in order to consider more realistic scenarios for clinical applications. 
Therefore, we model $S_a$ through a $\text{NN}_{S_a}$ rather than with a constant. 
This test aims to verify whether the PINN can correctly identify that $S_a$ is indeed constant, leading the neural network to predict an output that is nearly constant.
%In order to consider more realistic scenarios for clinical applications, we replace a scalar parameter $S_a$ with the active stress network $\text{NN}_{S_a}$, considering as a ground truth a spatially constant active contractility $S_a= \SI{118,08}{\kilo \pascal}$. 
As shown in~\Cref{tab:TI_hom_field}, the reconstruction of the parameter is very robust also in presence of noise with $LD = 0.05$, given the increased representational capacity of the model.
%and the network is able to detect that the field is homogenoeus, even w
% Mephisto:
% workstation_kfu/Ambrosi_SNR0_BL_SaField
% invSanetwT1.0e+01num_bcn_scales1_1_1_1_1_weight_bcn_scales8000_5_5_50_100_numPDE2048numBCN96numData500adam31000bfgs350000wPDE1.0e+04wFit1.0e+08wBCN1.0e+01wT1.0e-08wM1.0e-02wS0.0e+00seed{s}L32_16_8_BLRWFSNR0.0e+00
%
% Ambrosi_SNR5_BL_SaField
% invSanetwT1.0e+02num_bcn_scales1_1_1_1_1_weight_bcn_scales8000_5_5_50_100_numPDE8192numBCN512numData500adam31000bfgs350000wPDE1.0e+05wFit5.0e+07wBCN1.0e+02wT1.0e-08wM1.0e-02seed{s}BLSNR5.0e-02
\begin{table}[ht]
\centering
\begin{tabular}{cccc} 
\toprule
\textbf{$LD$} & \textbf{no. seeds} & \textbf{$L^2$ rel. err. on $u$} & \textbf{$\epsilon_{S_a; rel}$}\\
\midrule
0.00 & 10 & \num{1.62e-02} & \num{5.26e-02} \\ \midrule
0.05 & 10 & \num{4.01e-02} & \num{3.60e-02} \\
\bottomrule
\end{tabular}
\caption{
Performance for the quasi-static case, with $S_a$ modelled as a neural network. The first column indicates the noise level, the second column the number of successful seeds. The test case is repeated with ten seeds in total. The third column shows $\| \text{NN}_{\bu} - \bu \|/\|\bu \|$, the $L^2$ relative error for the state $\bu$ evaluated on the ground-truth mesh. The last column shows the $L^1$ relative error on $S_a$ w.r.t. the ground-truth value, evaluated on \num{1000} randomly chosen points. Both errors are first averaged over the different seeds and then averaged over the values from \num{9900} to \num{10000} iterations.
}
\label{tab:TI_hom_field}
\end{table}
\subsubsection{Time-dependent model}
In this section we consider the full time-dependent model of Eq.~\eqref{eq:pde}.
Parameter estimation in time-dependent cardiac models presents additional computational challenges for PINNs, as the network must accurately capture both spatial and temporal dynamics simultaneously, requiring careful balancing of loss terms to prevent the temporal derivatives from dominating the optimisation landscape while ensuring consistent convergence across the full spatio-temporal domain.
We include the time variable $t$ as an additional input of the neural network, and the aim of the following examples is to reconstruct the time-dependent displacement field and the parameter  $\sigma_0$ in Eq.~\eqref{eqn:bestel-clement-sorine}.
We consider a constant ground-truth parameter $\sigma_0$ that is modelled both as a scalar trainable parameter and as a field.
Since the active stress amplitude is given by $\sigma_0(\bx)S_a^1(t)$ and hence decouples space and time, we only need one solution of the time trajectory $S_a^1(t)$. 
This solution is calculated using standard ODE solvers. 
In particular, we use SciPy's implementation of the Radau method, an implicit fifth-order Runge-Kutta method.
The obtained discrete data is then interpolated using a second neural network. 
However, note that this is a pre-processing step. 
The obtained network $\mathrm{NN}_{S_a^1}(t)$ is not trainable afterwards.
For the following test cases, the data source consists of displacement data on \num{10000} randomly chosen points sliced on 5 parallel planes orthogonal the $y$-axis, with additional \num{100} points for the initial condition, where $t=\SI{160}{\milli \second}$ serves as an implicitly given initial condition. 
This amounts to approximately 53 data points per \si{\milli \second} since the considered time interval ranges from \SIrange{160}{350}{\milli \second}, corresponding to a systolic phase in the cardiac cycle.
\paragraph{Homogeneous parameter as a scalar trainable parameter}
\begin{figure*}[ht]
    \centering
    % First subfigure
    \centering
    \includegraphics[width=\linewidth]{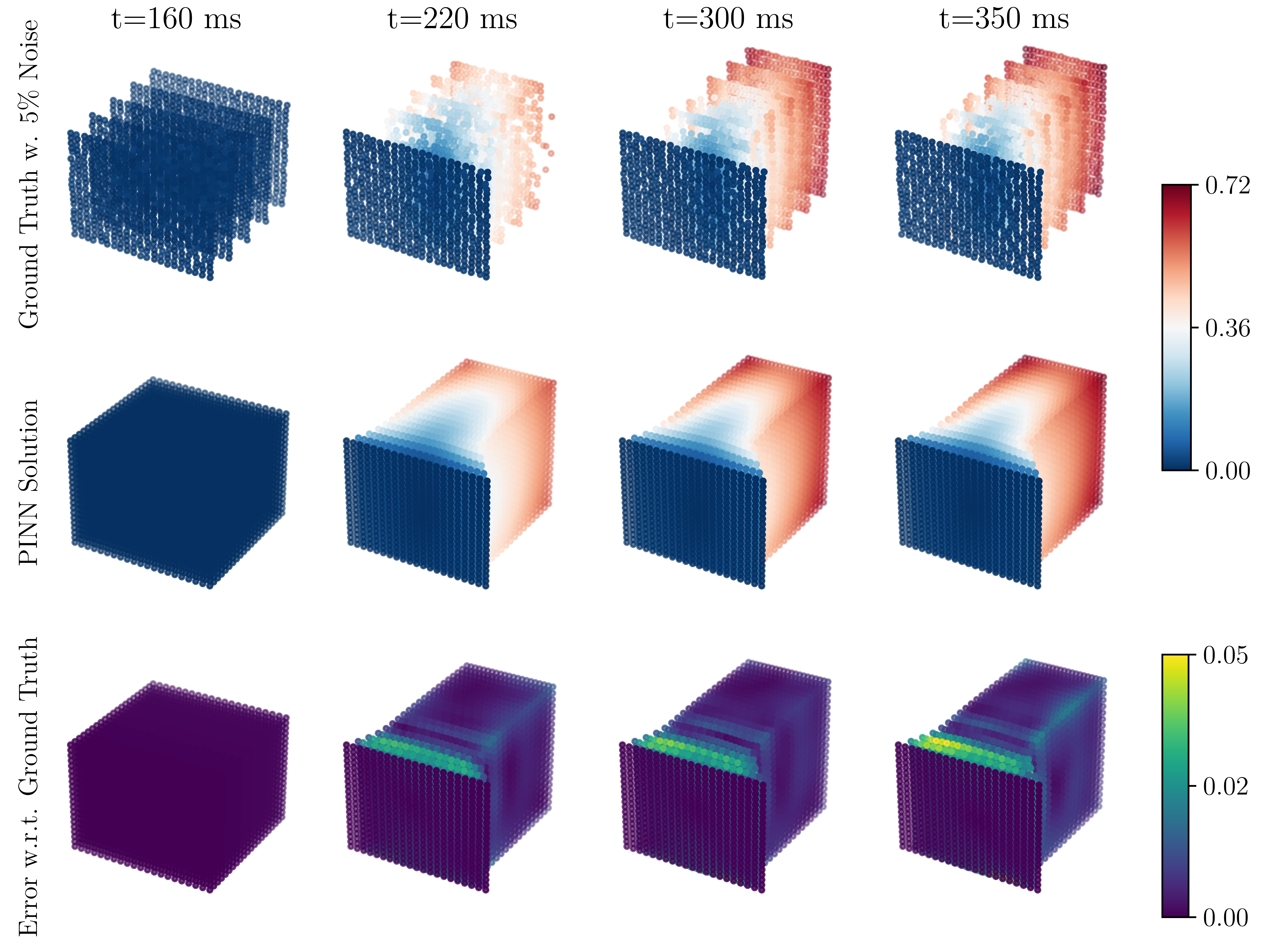}
    \label{fig:figure1}
    \caption{Time evolution of the PINN solution. Top row: ground-truth data corrupted with noise corresponding to $LD=0.05$ and then sliced along the $y$-axis. Middle row: PINN reconstruction of the displacement field. Bottom row: absolute error between the PINN reconstruction and the noise-free ground-truth solution.}
    \label{fig:time_dep_scalar}
\end{figure*}
In this test case we model the unknown parameter $\sigma_0$ as a scalar quantity (constant in space and time).  
We consider for the training \num{1000} Adam epochs followed by a maximum amount of \num{40000} BFGS epochs. 
However, the BFGS iterations stopped earlier for every instance leading to an average amount of \num{16770} and \num{27040} BFGS epochs for the noiseless case and the noisy case with $LD=0.05$ respectively. 
The time evolution of the ground-truth solution affected with noise, the PINN solution and the absolute error between the PINN reconstruction and the noiseless ground truth are given in~\Cref{fig:time_dep_scalar}.
It is possible to note that the error concentrates close to the Dirichlet boundary and the other boundary edges. 
% Mephisto: vsc5/B160_350_SNR0_BL_Slicing
% Mephisto: vsc5/B160_350_SNR5_BL_Slicing
\begin{table}[ht]
\centering
\begin{tabular}{cccc} 
\toprule
\textbf{$LD$} & \textbf{no. seeds} & \textbf{$L^2$ rel. err. on $\bu$} & \textbf{$\epsilon_{\sigma_0; rel}$}\\
\midrule
 0.00 & 7 & \num{3.49e-02} & \num{6.10e-02} \\ \midrule
 %&&&  \\ \midrule
 0.05 & 6 & \num{5.33e-02} & \num{6.48e-03} \\
\bottomrule
\end{tabular}
\caption{Performance for the time-dependent case, with $\sigma_0$ modelled as a single parameter. The first column indicates the noise level, the second column the number of successful seeds. The test case is repeated with ten seeds in total. The third column shows $\| \text{NN}_{\bu} - \bu \|/\|\bu \|$, the $L^2$ relative error for the state $\bu$ evaluated on the ground-truth mesh. The last column shows the relative error on $\sigma_0$ w.r.t. the ground-truth value. Both errors are first averaged over the different seeds and then averaged over the values from \num{19900} to \num{20000} iterations.
}
\label{tab:time_dep_scalar}
\end{table}
For the sake of completeness, we also report in~\Cref{tab:time_dep_scalar} the relative $L^2$ testing error for the state $\bu$ and the relative error of $\sigma_0$ w.r.t. the ground-truth value.
%, which are below 1\% and 5\%, respectively, also using scattered, noisy data.
%
\paragraph{Homogeneous parameter as a field}
% vsc5/B160_350_SNR5_sigma0Field
% invsignetwT1.0e+02weight_bcn_scales50000_500_500_1000_1000_numPDE30000numBCN5000numData10000adam1800bfgs1200adam31000bfgs320000wPDE1.0e-03wFit1.0e+08wBCD1.0e+06wBCN1.0e+01wIC0.0e+00seed{s}BLSNR0.0e+00
% vsc5/B160_350_SNR5_sigma0Field
% invsignetwT1.0e+02weight_bcn_scales50000_500_500_1000_1000_numPDE30000numBCN5000numData10000adam1800bfgs1200adam31000bfgs350000wPDE1.0e-03wFit1.0e+08wBCD1.0e+06wBCN1.0e+01wIC0.0e+00seed{s}BLSNR5.0e-02
As a second test case for the time-dependent scenario we model the unknown parameter $\sigma_0$ as a field (constant in time), given by a second neural network $\text{NN}_{\sigma_0}$ that is trained simultaneously to $\text{NN}_{\bu}$, with three internal layers of 12, 8, and 4 neurons, respectively.
The training is based on \num{1000} Adam epochs followed by a maximum amount of \num{20000} BFGS epochs. 
%However, the BFGS iterations stopped earlier for every instance leading to an average amount of \textcolor{Matthias}{still computing - to add here} and \num{10000} BFGS epochs for the noiseless case and the noisy case with $LD=0.05$, respectively. 
As illustrated in~\Cref{tab:time_dep_NN_field}, the model is able to accurately reconstruct the displacement field and estimate the ground-truth value of the parameter $\sigma_0$ also in presence of noise on the data.

\begin{table}[ht]
\centering
\begin{tabular}{cccc} 
\toprule
\textbf{$LD$} & \textbf{no. seeds} & \textbf{$L^2$ rel. err. on $u$} & $\epsilon_{\sigma_0 \text{; rel}}$\\
\midrule
% \num{0.00} & 9 & \num{9.65e-04} & \num{3.98e-02} \\ \midrule
 0.00 & 8 & \num{2.32e-02}& \num{2.23e-02}\\ 
 \midrule
 \num{0.05} & 9 & \num{2.56e-02} & \num{1.61e-02} \\
\bottomrule
\end{tabular}
\caption{
Performance for the time-dependent case, with $\sigma_0$ modelled as a neural network. The first column indicates the noise level, the second column the number of successful seeds. The test case is repeated with ten seeds in total. The third column shows $\| \text{NN}_{\bu} - \bu \|/\|\bu \|$, the $L^2$ relative error for the state $\bu$ evaluated on the ground-truth mesh. The last column shows the $L^1$ relative error on $\sigma_0$ w.r.t. the ground-truth value, evaluated on \num{10000} randomly chosen points. Both errors are first averaged over the different seeds and then averaged over the values from \num{19900} to \num{20000} iterations.
}
\label{tab:time_dep_NN_field}
\end{table}
%
%
%\begin{figure*}[htbp]
%    \centering
%    % First subfigure
%    \begin{subfigure}[b]{0.45\textwidth}
%        \centering
%        \includegraphics[width=\textwidth]{invSanetwT1.0e+01numPDE2048numBCN96numData500wPDE1.0e+04wFit1.0e+08wBCN1.0e+01wT1.0e-08wM1.0e-02wS0.0e+00_RWFcomparison.pdf}
%        \caption{Without noise}
%        \label{fig:figure1}
%    \end{subfigure}
%    % Second subfigure
%    \begin{subfigure}[b]{0.45\textwidth}
%        \centering
%        \includegraphics[width=\textwidth]%{invSanetwT1.0e+02numPDE8192numBCN512numData500wPDE1.0e+05wFit5.0e+07wBCN1.0e+02wT1.0e-08wM1.0e-02comparison.pdf}
%        \caption{With 5\% noise}
%        \label{fig:figure2}
%    \end{subfigure}
%    \caption{Comparison of the reconstruction for active stress field with 5 seeds.}
%\end{figure*}

\subsection{Heterogeneous test case - detection of scars}
\label{sec:res_heterogeneous}
For the heterogeneous test cases, we consider again a quasi-static approximation. 
The objective of the following examples is to test the ability of the method to properly reconstruct heterogeneous contractility fields and detect the presence of scars in the cardiac tissue.
The displacement network has a rectangular architecture with 3 layers with a width of 30 neurons. 
Additionally, a residual connection from input to output layer is considered in order to mitigate vanishing gradients.
The active stress network incorporates a Fourier feature embedding with 
%feature dimension equal to 24 and 
Fourier feature frequencies sampled from a zero-mean normal distribution with $\sigma_F=3$. 
This layer is followed by three internal layers of 12, 8, and 4 neurons, respectively. 
For the results shown in this section, the slope parameter of the sigmoid function in the output layer is set to $\alpha=8$. 
However, the results are robust for $\alpha \in [1,10]$.
In addition, the final output is restricted by physiological interval constraints with a maximal value of \SI{120}{\kilo\pascal} and a minimal value of \SI{0.1}{\kilo\pascal}.
The second step of the optimisation procedure is run with 1000 epochs of ADAM followed by 40000 epochs of BFGS.
Note that for these test cases we also use strain data uniformly sampled on random locations to achieve a better identifiability of the active stress parameter $S_a$.
In addition to reconstructing the active stress field, our ultimate goal is to identify fibrotic regions. 
However, due to the lack of assumptions regarding topological properties — such as the number of regions and their connectivity — we apply a simple thresholding approach.
As detailed in Appendix~\ref{sec:appendix_threshold}, the optimal threshold value is approximately \SI{50.00}{\kilo\pascal}, which is selected as the default value.
\subsubsection{Single spherical scar including a grey zone}
\label{sec:one_scar}
\begin{figure}[h!]
\centering
    \includegraphics[width=1.\linewidth]{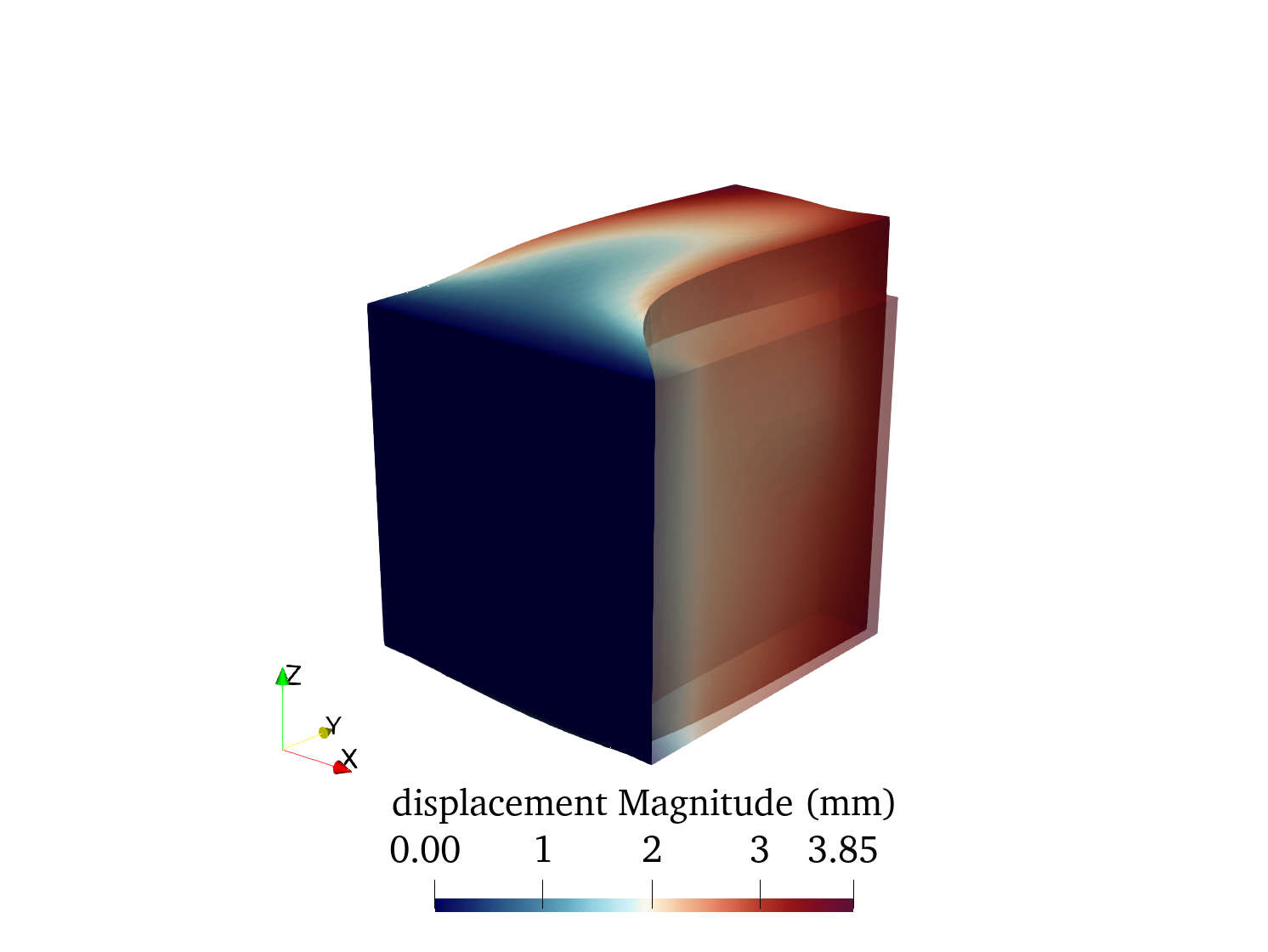}
    \caption{One-scar test case. Ground-truth FEM solution of the displacement field. The light-shaded cube represents the reference configuration, whereas the coloured object is in deformed configuration. %Colour coding is according to the displacement magnitude.
    }
    \label{fig:1scar}
\end{figure}

\begin{figure*}[ht]
    \centering
    %old one:\includegraphics[width=0.8\textwidth]{f1_Ambrosi_SNR5_BL_Scar_grey_big_RBA_PDE.pdf}
    %new calculation: (f1_strnumData9000invSanetfousig3.0e+00invSanetL12_12_8_4_invSasiga8.0e+00invSanetsigmoidinvSasigmin1.0e-01invSanetwTV1.0e-04invSanetwT0.0e+00numBCN1024numData9000adam31000seed4SNR5.0e-02)
    \includegraphics[width=0.8\textwidth]{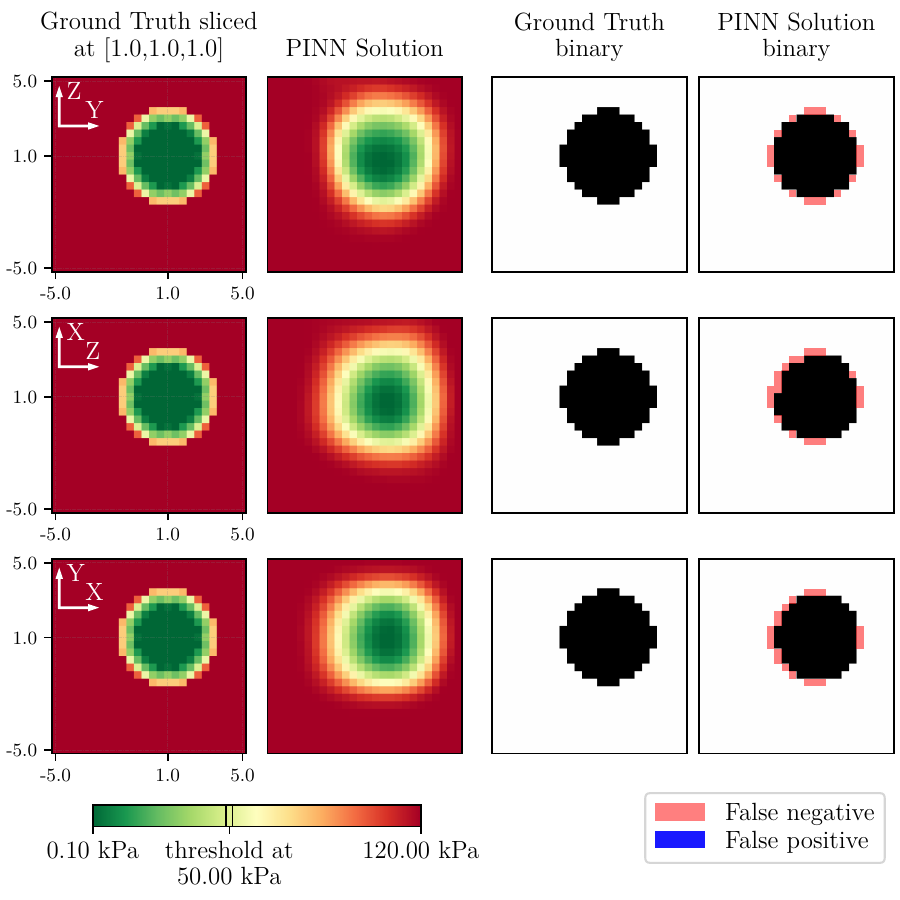}
    \caption{Active stress field reconstruction for a case with a central scar surrounded by a grey zone (border zone). From left to right: (1) Ground-truth solution showing the spatial distribution of active stress parameter $S_a$ with values ranging from 7.87 kPa (scar core) to 118.08 kPa (healthy tissue); (2) PINN reconstruction of $\mathrm{NN}_{S_a}$ showing the estimated active stress field; (3) Binary classification of the ground-truth after thresholding at 50 kPa, with black regions indicating tissue classified as scarred; (4) Binary classification of the PINN reconstruction after applying the same threshold, where red areas indicate false negatives (scarred tissue incorrectly classified as healthy) and blue areas indicate false positives (healthy tissue incorrectly classified as scarred).}
    \label{fig:1scar_rec}
\end{figure*}

% For parameter values see Mephisto:
% vsc5/Ambrosi_SNR5_BL_Scar_grey_big_RBA_PDE/Data/invSanetfousig3.0e+00invSanetL12_12_8_4_invSasiga8.0e+00invSanetsigmoidinvSasigmin1.0e-01invSanetwTV1.0e-04bcweight_bcn_scales1_0_0_1_1_invSanetwT0.0e+00numPDE16000numBCN1024wPDE1.0e-01wFit1.0e+01wBCN1.0e-03wS1.0e+01seed2SNR1.0e-01.json
This test case features a fibrotic scar region surrounded by a border zone (grey zone) characterised by reduced contractility relative to healthy myocardium, but with tissue properties not yet indicating complete fibrosis, corresponding to the ground-truth solution depicted in~\Cref{fig:1scar}. Denoting with $\mathbf c=(1.0,1.0,1.0)^T \si{\milli \meter}$ the centre, the ground-truth active stress parameter field is given by
\[
S_a(\bx) = \begin{cases}
    \SI{7.87}{\kilo \pascal}, & \text{ for } \norm{\bx- \mathbf c}<1.9, \\
    \SI{37.79}{\kilo \pascal}, & \text{ for } 1.9 \leq \norm{\bx- \mathbf c}<2.5 \\
    \SI{118.08}{\kilo \pascal}, & \text{ elsewhere}.
\end{cases}
\]
%\textcolor{Matthias}{to do: consider case with grey zone at 30\% of healthy contractility if possible (and if you get better results)}
\noindent
Both displacement and strain data used for training are corrupted with noise corresponding to $LD=0.05$. 
The active stress network incorporates Fourier features with 
%a frequency of $\sigma_F=3$ and 
an output feature space of dimension 24.
%, followed by a trapezoidal architecture with internal layers of sizes 12, 8, 4 neurons, respectively. 
\begin{table}[ht]
\centering
\begin{tabular}{cc} 
\toprule
\multicolumn{2}{c}{\textbf{Hyperparameter values}}\\
\midrule
displacement data points & \num{9000} \\ \midrule
strain data points & \num{9000} \\ \midrule
collocation points for PDE & \num{16000} \\ \midrule
collocation points for BC & \num{5120} \\ \midrule
data weight &  \num{1e2} \\ \midrule
PDE weight &  \num{1e-1} \\ \midrule
BC weight & \numrange{1e-4}{1e-3} \\ \midrule
$\text{NN}_{S_a}$ regularisation & \num{1e-4} \\
\bottomrule
\end{tabular}
\caption{Single scar test case.}
\label{tab:TI_1scar_param}
\end{table}
The parameters considered in this test case are summarised in~\Cref{tab:TI_1scar_param}.
%The overall network is then a mapping $\R^3 \rightarrow \R^{24} \rightarrow \R^{12} \rightarrow \R^8 \rightarrow \R^4 \rightarrow \R$.
The model performance for this test case, as shown in \Cref{fig:1scar_rec}, is very satisfactory in reconstructing the active stress parameter.
However, the reconstructed parameter field appears smoother than the ground-truth data, attributable to the well-known phenomenon of spectral bias. 
The reconstructed image with threshold reveals a slight underestimation of the scarred area by our method, influenced by the presence of the grey zone and the effect of spectral bias. 
Nevertheless, it's worth noting the high accuracy of the reconstruction, with minimal misclassification, even when utilising only limited strain and displacement data.
%
% For the sake of completeness, we also show in~\Cref{fig:1scar_strain} the performance of the method in reconstructing the Green-Lagrange strain tensor. 
% \begin{figure*}[ht]
%     \centering
%     \includegraphics[width=0.8\textwidth]{Ambrosi_SNR5_BL_Scar_grey_RBA_PDE_strain.pdf}
%     \caption{Reconstruction of the Green-Lagrange strain for the single scar test case. From left to right: ground-truth Green-Lagrange strain with $LD=0.1$, PINN reconstruction based on the displacement network $\text{NN}_{\bu}$, difference between the ground truth without noise and the PINN reconstruction.}
%     \label{fig:1scar_strain}
% \end{figure*}
%
% \subsubsection{One ellipsoidal scar}
% \textcolor{Federica}{see if we get nice results!}
\subsubsection{Two spherical scars}
\label{sec:two_scar}
\begin{figure}[h!]
\centering
    \includegraphics[width=1.\linewidth]{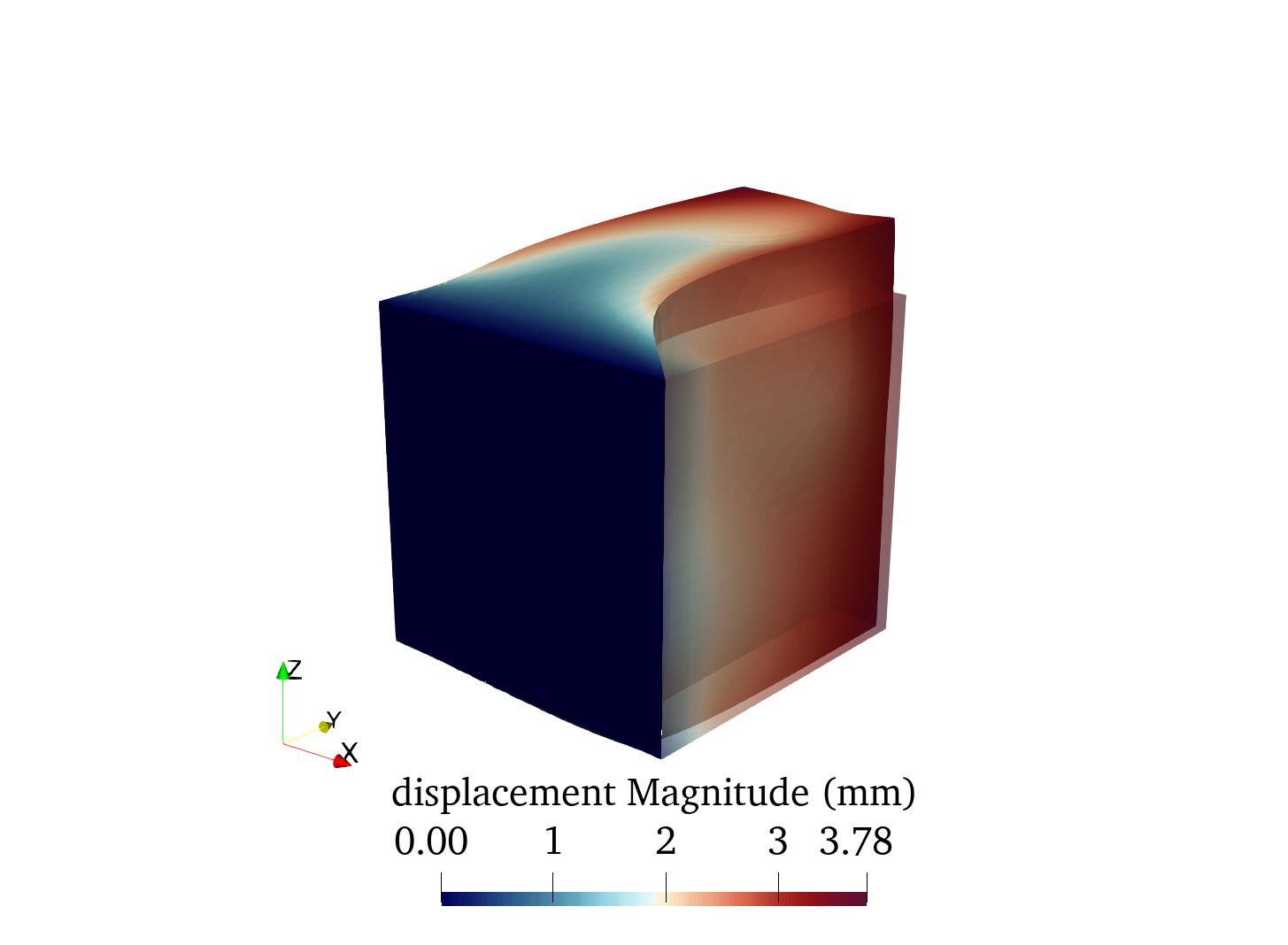}
    \caption{Two-scar test case. Ground-truth FEM solution of the displacement field. The light-shaded cube represents the reference configuration, whereas the coloured object is in deformed configuration. %Colour coding is according to the displacement magnitude.
    }
    \label{fig:2scars}
\end{figure}

\begin{figure*}[ht]
    \centering
    \includegraphics[width=0.8\textwidth]{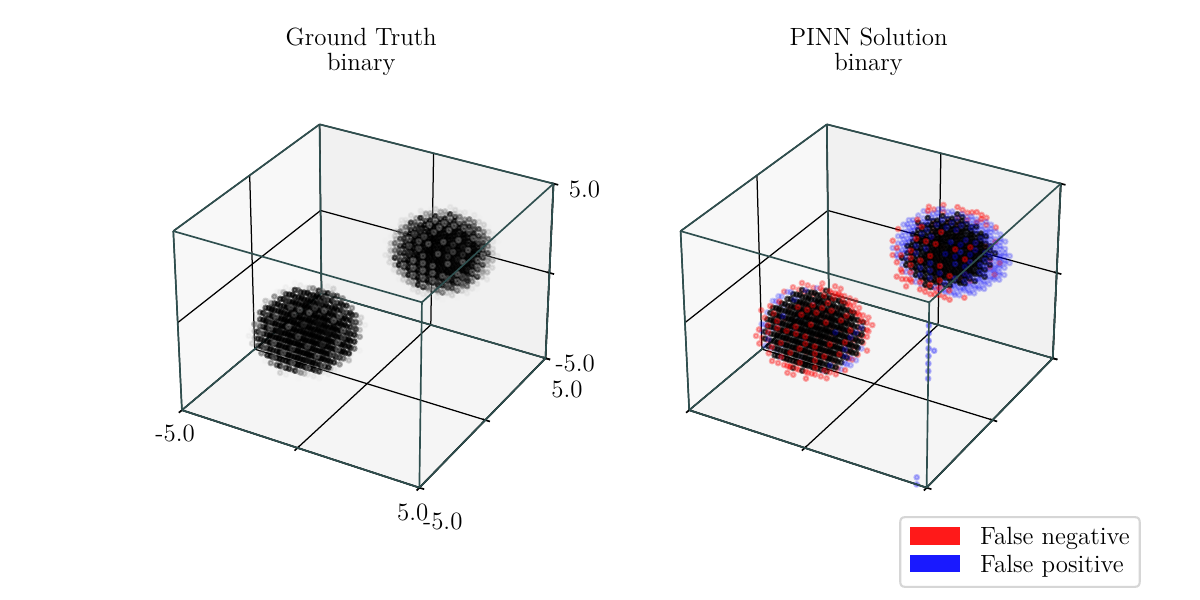}
    \caption{Two-scar test case. The plots show the thresholded active stress field including two scars. Left: ground-truth solution. Right: PINN reconstruction of $\text{NN}_{S_a}$. Red areas indicate false negatives (scarred tissue incorrectly classified as healthy) and blue areas indicate false positives (healthy tissue incorrectly classified as scarred).}
    \label{fig:2scar_rec}
\end{figure*}

% For parameter values see Mephisto:
%vsc5/Ambrosi_SNR5_BL_2Scar_shift_RBA_PDE/Data/invSanetfousig3.0e+00invSanetL12_12_8_4_invSasiga8.0e+00invSanetsigmoidinvSasigmin1.0e-01invSanetwTV1.0e-03bcweight_bcn_scales1_0_0_1_1_invSanetwT0.0e+00numPDE16000numBCN1024wPDE1.0e+01wFit1.0e+03wBCN1.0e-01wS1.0e+03seed2SNR5.0e-02.json
For this test case, two scars are present in the ground-truth data, shown in~\Cref{fig:2scars}.
Strain and displacement data are used, both corrupted with noise corresponding to an $LD$ value of \num{0.05}. 
The field network $\text{NN}_{S_a}$ includes Fourier features with
%a frequency of $\sigma_F=3$ and features 
output dimension of \num{18}. 
%The output of the network is followed by box constraints with scaling values $S_{a,\max}=\SI{120}{\kilo\pascal}$ and $S_{a, \min}=\SI{0.1}{\kilo\pascal}$. 
\Cref{tab:TI_2scar_param} summarises the parameters considered for this test case.
\begin{table}[ht]
\centering
\begin{tabular}{cc} 
\toprule
\multicolumn{2}{c}{\textbf{Hyperparameter values}}\\
\midrule
displacement data points & \num{12000} \\ \midrule
strain data points & \num{12000} \\ \midrule
collocation points for PDE & \num{16000} \\ \midrule
collocation points for BC & \num{5120} \\ \midrule
data weight &  \num{1e3} \\ \midrule
PDE weight &  \num{1e1} \\ \midrule
BC weight & \numrange{1e-2}{1e-1} \\ \midrule
$\text{NN}_{S_a}$ regularisation & \num{1e-3} \\
\bottomrule
\end{tabular}
\caption{Two-scar test case.}
\label{tab:TI_2scar_param}
\end{table}
%
%The relative contributions of the loss weights (i.e. normalised by the sum) is given in~\Cref{tab:TI_2scar_weight}.
%\begin{table}[ht]
%\centering
%\begin{tabular}{cc} 
%\toprule
%\multicolumn{2}{c}{\textbf{Hyperparameter values}}\\
%\midrule
%data weight & \num{0.990000} \\ \midrule
%PDE weight &  \num{0.009900} \\ \midrule
%BC weight &   \num{0.000099} \\
%\bottomrule
%\end{tabular}
%\caption{Two-scar test case.}
%\label{tab:TI_2scar_weight}
%\end{table}
%Figure~\ref{fig:2scar_prediction} depicts the performance %of the algorithm for this test case.
\Cref{fig:2scar_rec} depicts the PINN reconstructions of the two scars after applying a threshold of \SI{50}{\kilo\pascal}.
Also in this test case, the estimation of both scars and their contours is very satisfactory.

\section{Discussion}
\label{sec:Discussion}
% Add this table to the Discussion section (from GPT)
% \begin{table*}[h]
% \centering
% \caption{Comparison of our PINN-based approach with existing methods for active property estimation}
% \begin{tabular}{p{3.5cm}p{2.5cm}p{2.5cm}p{2.5cm}p{2.5cm}}
% \toprule
% \textbf{Method} & \textbf{Spatial Resolution} & \textbf{Data Requirements} & \textbf{Computational Cost} & \textbf{Detection of Heterogeneities} \\
% \midrule
% PDE-constrained optimization \cite{pozzi2024reconstruction} & High & Displacement only & High (multiple FEM solves) & Yes, with prior assumptions \\
% \\
% Data assimilation \cite{imperiale2021sequential} & Medium (AHA regions) & Tagged-MRI & Medium & Limited to predefined regions \\
% \\
% Tikhonov regularization \cite{kovacheva_estimating_2021} & Medium & Wall motion & Medium-High & Yes, with smoothness priors \\
% \\
% Our PINN approach & High & Displacement and strain & Low-Medium (single training) & Yes, without prior assumptions \\
% \bottomrule
% \end{tabular}
% \label{tab:method_comparison}
% \end{table*}

The findings of this study suggest that our PINN-based approach is effective for estimating active contractility properties in the context of soft tissue nonlinear biomechanics. 
It is capable of effectively utilising sparse, limited, and noisy data while considering heterogeneous parameter fields and time-dependent PDEs. 
We remark that for the estimation of a heterogeneous parameter field we also make use of strain data to achieve satisfactory results. 
However, this does not represent a large limitation of the method, since such data can be accessed in clinical contexts, f.e. with Tagged-MRI, or derived in a post-processing step from displacement data~\cite{lopez2023warppinn, verzhbinsky2019estimating}.
Notably, our problem formulation and training strategy allow for a substantially reduced number of neurons, epochs, and training points compared to previous works utilising PINNs for inverse problems in elasticity~\cite{Haghighat2021a,kamali2023elasticity}. 
This is achieved despite the inclusion of time-dependent nonlinear mechanics, anisotropic constitutive laws, and a three-dimensional framework in our study.
Preliminary results indicate that a three fully connected hidden layers setup with 32, 16, and 8 neurons for $\mathrm{NN}_{\bu}$ represents a favourable compromise between network representation capacity, computational costs, and the prevention of overfitting for constant active stress parameters. 
For scenarios involving heterogeneous parameter fields, we employ a three-layer architecture with 30 neurons each for the displacement network $\mathrm{NN}_{\bu}$, and three layers with 12, 8, and 4 neurons for $\mathrm{NN}_{S_a}$ or $\mathrm{NN}_{\sigma_0}$, for the quasi-static or time-dependent cases, respectively.
In terms of the NN optimisers, we utilise a combination of first- and second-order optimisers, a commonly adopted approach in complex optimisation routines to enhance estimation accuracy~\cite{regazzoni2021physics} and more recently in the context of PINNs~\cite{rathore2024challenges, caforio2024physics}. 
%
%Additionally, we implemented a two-step approac, where the minimisation of the physics-informed loss function is preceded by the minimisation of a reduced loss function containing the data fidelity term only. 
To mitigate the effects of spectral bias in heterogeneous test cases, we employ Fourier Features solely in the parameter neural network.
To get more insights on the most favourable combinations of loss weights to achieve the best estimation of the active stress parameters, we perform a thorough analysis of the apparent Pareto front, detailed in~\Cref{sec:pareto_front}.
We also analyse the identifiability of the active stress parameters on the boundary in Appendix~\ref{sec:appendix_identifiability} in relation to the anisotropy of the tissue. 
In addition, in this work we improve the method proposed in~\cite{caforio2024physics} and add adaptive weighting schemes and residual-based attention mechanisms (as illustrated in~\Cref{sec:adaptive_weights},~\Cref{sec:RBA} and Appendix~\ref{sec:appendix_adaptive_weights}), include \textit{ad hoc} regularisation terms to cope with identifiability issues associated with the specific choice of boundary conditions (as discussed in~\Cref{sec:reg} and Appendix ~\ref{sec:appendix_reg}).
Finally, we exactly enforce Dirichlet boundary conditions in the NN architecture (as shown in~\Cref{sec:exact_BCD} and Appendix~\ref{sec:appendix_BC_weak}) and compare these results with the PINN performance considering Robin BC that yield similar mechanic deformation in the ground-truth solution (Appendix~\ref{sec:appendix_BC_robin}), also testing the robustness of the prediction in presence of BC misspecification. 
Our study also involves training an ensemble of neural networks with different initial weights and biases to measure the reliability of the model and the robustness with respect to random initialisation. 
Regarding the initial guess for the estimated (constant) parameter, we consider at least a 50\% overestimation, a reasonable choice considering the physiological range of the parameters. 
We also make the assumption that the tissue's constitutive law is known \emph{a priori}, based on existing comparative works demonstrating the ability of different constitutive laws, under specific parametrisation, to match the same end-diastolic-pressure-volume-relationship.
We previously showed in~\cite{caforio2024physics} that the different constitutive laws can provide similar displacement fields and that the PINN method is robust with respect to this source uncertainty. 
%Model uncertainties can be structurally embedded in the PINN estimation, as in~\cite{zou2023correcting}, f.e. in combination with Bayesian PINNs (B-PINNs) \cite{yang2021b} 
% We identified three potential applications of this method: 1) calibrating patient-specific, heterogeneous cardiac biomechanical models based on clinical data, 2) deriving relative, qualitative information on the spatial variation of tissue material properties to identify pathological tissue, and 3) using the PINN to evaluate different constitutive laws by computing the residual of equations through automatic differentiation.
Furthermore, we test the method for evaluating tissue heterogeneities and non-invasively detecting scar regions without prior assumptions about the shape of scars and in the absence of stress data. 
The ability of our method to accurately identify and characterise the scar region, including its surrounding grey zone, has important clinical implications. 
In post-infarction patients, precisely delineating the boundary between the scar core and the potentially salvageable grey zone tissue is crucial for therapy planning and risk stratification. 
The slight underestimation of the scar area observed in our results is consistent with the challenge of detecting partially viable tissue in the grey zone, which also occurs in clinical imaging methods like late gadolinium enhancement MRI~\cite{fahmy2021improved}. 
Notably, our method achieves this distinction using only displacement and strain data, without requiring contrast agents or additional imaging sequences, potentially simplifying clinical workflows while providing comparable diagnostic information.
As future steps we plan the incorporation of more realistic and representative geometries~\cite{costabal2024delta}, such as patient-specific computational domains derived from the segmentation of clinical images, with realistic fibre distributions.
%The sampling of complex geometries in PINNs can be performed,for example, using signed distance functions.
In addition, for this work, we consider an idealised test case with constant cardiac fibre orientation. 
%Future studies will include realistic fibre distributions.
%, leveraging the authors' extensive expertise in this domain.

%add early stopping criterion? 

\section{Conclusion}
\label{sec:Conclusion}
This study presents a significant advancement in the accurate reconstruction of displacement fields and estimation of active cardiac material properties by developing a PINN methodology specifically tailored to time-dependent biomechanical models.
The proposed training builds upon the previously suggested method in~\cite{caforio2024physics} and contributes in several key technical aspects:
First, our specially designed network architecture and ad-hoc selection of optimisers enables efficient learning, significantly reducing computational requirements compared to traditional PINN methods.
In particular, the use of Fourier feature embeddings specifically for the parameter network addresses the spectral bias problem, improving detection of high-frequency features crucial for identifying scar boundaries.
In addition, the proposed thresholding methodology enables scar detection without prior shape assumptions.
Second, employing adapted regularisation strategies, along with the exact imposition of Dirichlet boundary condition and modified boundary loss terms, notably improves parameter identifiability near boundaries -- a challenge in active stress estimation due to the anisotropic nature of cardiac tissue. 
%This allows accurate parameter reconstruction throughout the entire domain without requiring uniform observability.
Third, integrating residual-based attention mechanisms and adaptive weight balancing enhances convergence properties, achieving accurate estimation with fewer training points and epochs than comparable PINN approaches for inverse elasticity problems, despite the added complexity of nonlinear cardiac mechanics and time-dependent dynamics.
Fourth, our approach to time-dependent parameter estimation with decoupled space-time representations results in a drastic reduction in computational cost for the estimation.
A comprehensive analysis of the apparent Pareto Front for the scalar parameter estimation test case provides better insight into optimal combinations of loss weights in the resulting multi-objective optimisation problem.
With the aid of this novel methodology, it is possible to estimate a spatial parameter based solely on a limited number of displacement and, in clinically-relevant cases, strain measurements, circumventing the requirement for stress data, which are typically inaccessible within a realistic clinical setting, particularly in cardiac applications.
%
%The accuracy and robustness of the method were proved in various test scenarios representing both healthy and pathological states. 
The proposed methodology could potentially contribute to improved risk stratification for cardiac pathologies associated with the presence of fibrotic tissue, f.e., by enabling the non-invasive characterisation of infarct regions, the quantification of contractile impairment in cardiomyopathies, or therapy response monitoring.
%by identifying regions of impaired contractility that may benefit from intervention. 
These applications represent promising directions for translating our computational approach into tangible clinical benefits.

\section*{Acknowledgments}
The authors acknowledge Prof. Alfio Quarteroni (Politecnico di Milano) for his valuable insights on the project and Dr. Matthias Gsell (Medical University of Graz, Austria) for his technical support with the 3D FEM biomechanical model.
FEM simulations for this study were performed on the Vienna Scientific Cluster (VSC-4, VSC-5) under PRACE projects \#71962 \#72420, which is maintained by the VSC Research Center in collaboration with the Information Technology Solutions of TU Wien. 
FC, FR and SP are members of the INdAM research group GNCS.
FC acknowledges support from L'Oréal UNESCO Österreich and the Land Steiermark UFO Grant No. 3026.
EK acknowledges support from the BioTechMed-Graz Young Researcher Grant \enquote{CICLOPS ---  Computational Inference of Clinical Biomarkers from Non-Invasive Partial Data Sources}.
This research was funded in whole or in part by the Austrian Science Fund (FWF)  under grant 10.55776/P37063     to CMA.
FR and SP have received support from the project PRIN2022, MUR, Italy, 2023-2025, P2022N5ZNP “SIDDMs: shape-informed data-driven models for parametrized PDEs, with application to computational cardiology”, funded by the European Union (Next Generation EU, Mission 4 Component 2).
FR and SP acknowledge the support by the MUR, Italian Ministry of University and Research (Italy), grant Dipartimento di Eccellenza 2023-2027.

\bibliographystyle{unsrtnat}
\bibliography{biblio}  %%% Uncomment this line and comment out the ``thebibliography'' section below to use the external .bib file (using bibtex) .

%%%%%%%%%% Merge with supplemental materials %%%%%%%%%%
%\pagebreak
\begin{center}
\textbf{\large Supplementary Materials}
\end{center}
%%%%%%%%%% Prefix a "S" to all equations, figures, tables and reset the counter %%%%%%%%%%
\setcounter{equation}{0}
\setcounter{figure}{0}
\setcounter{table}{0}
\setcounter{page}{1}
\setcounter{section}{0}
\makeatletter
\renewcommand{\thesection}{S-\arabic{section}}
\renewcommand{\theequation}{S\arabic{equation}}
\renewcommand{\thefigure}{S\arabic{figure}}
\renewcommand{\bibnumfmt}[1]{[S#1]}
\renewcommand{\citenumfont}[1]{S#1}
%%%%%%%%%% Prefix a "S" to all equations, figures, tables and reset the counter %%%%%%%%%%

\section{Identifiability of Active Stress on the Boundary}
\label{sec:appendix_identifiability}
The anisotropic nature of active stress can lead to further problems on identifiability on the boundary. Assume for simplicity that $\bf{f}_0=e_1$ i.e. that the fibre direction is aligned with the first unit vector. The active stress part is then given by the matrix

\begin{align*}
\mathbf{P}_{\mathrm{act}} &=S_a \frac{\mathbf{F} \mathbf{f}_0 \otimes \mathbf{f}_0}{\sqrt{\mathbf{F} \mathbf{f}_0 \cdot \mathbf{F} \mathbf{f}_0}} \\
&=
\frac{S_a}{\sqrt{1 + 2 \nabla \mathbf{u}_{1,1} + \sum_j \nabla \mathbf{u}_{j,1}^2}}
\begin{pmatrix}
    1+\nabla \mathbf{u}_{1,1} & 0 & 0 \\
    0 & 0 & 0 \\
    0 & 0 & 0
\end{pmatrix}.
\end{align*}
Hence, if we look at the Neumann boundary condition, which reads
\[
\left( \mathbf{P}_{\mathrm{pas}} + \mathbf{P}_{\mathrm{act}} \right) \mathbf{n}=0 \quad \text { on } \Gamma_N,
\]
we see that the active contribution vanishes on parts of the boundary where the normal vector $\bn$ is perpendicular to the fibre direction $\bf{f}_0$.\\
On the other side, if $\alpha \coloneqq \langle e_1, \bn \rangle \neq 0$, then we could directly invert for $S_a$ via
\begin{equation}\label{eqn:Saexplicit}
\begin{aligned}
S_a &= - \frac{
\sqrt{1 + 2 \nabla \mathbf{u}_{1,1} + \sum_j \nabla \mathbf{u}_{j,1}^2}
}{\alpha \left( 1+\nabla \mathbf{u}_{1,1} \right)} \langle \mathbf{P}_{pas; 1, \cdot},  \mathbf{n} \rangle,
\end{aligned}
\end{equation}
provided that we have correctly learned $\nabla \bu$.\\
In particular, if $\nabla \bu$ tends to vanish near the boundary, then so does $\mathbf{P}_{pas}$ and hence also $S_a$.
\section{Comparison of results using adaptive weights / RBA }
\label{sec:appendix_adaptive_weights}
\begin{figure*}[ht]
    \centering
\includegraphics[width=0.7\textwidth]{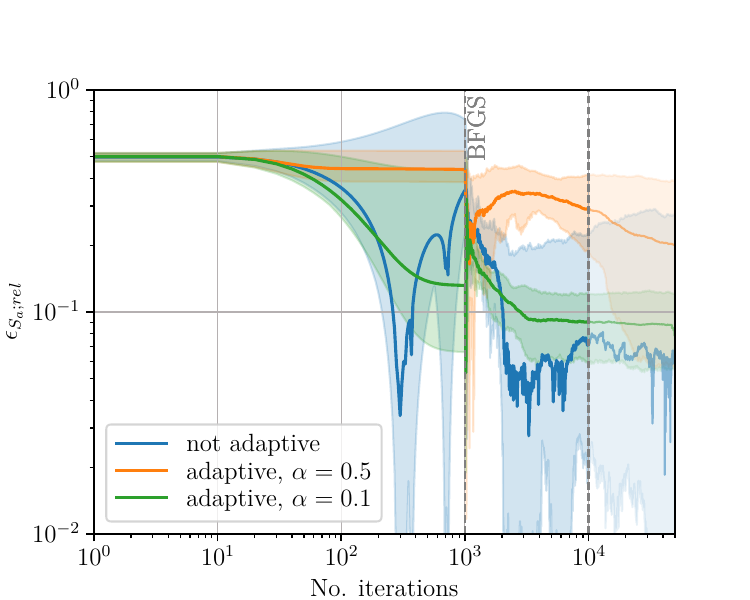}
    \caption{Comparison of the reconstructions of $S_a$ using weight balancing. The test case and hyperparameters are taken from \Cref{sec:res_static} with a noise level of $LD=0.05$. The plot shows trajectories of the relative error $\epsilon_{S_a;rel}$ where the blue line corresponds to the non-adaptive setting as already presented in \Cref{fig:T1_hom_single:subnoise}. The other two lines show the same setting, however using weight balancing for the loss terms. In particular, also the parameter $0<\alpha<1$ controlling the moving average is varied. Each setup is undertaken with ten different seeds. The solid line depicts the geometric mean over the seeds; the shaded region is the area spanned by the trajectories. The second dashed vertical line marks the selected end of training at 10k BFGS epochs, for which we report performance and associated errors. For completeness, we also display the algorithm's performance up to 50k epochs.}
    \label{fig:weight_balancing}
\end{figure*}

\begin{figure*}[ht]
    \centering
    \includegraphics[width=0.7\textwidth]{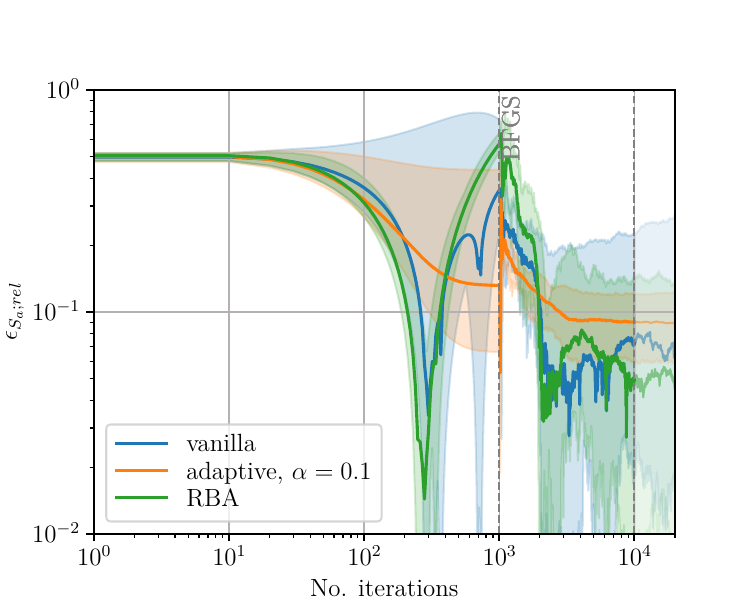}
    \caption{Comparison of the reconstructions of $S_a$ using different weighting schemes. The test case and hyperparameters are taken from \Cref{sec:res_static} with a noise level of $LD=0.05$. The plot shows trajectories of the relative error $\epsilon_{S_a;rel}$ where the blue line corresponds to the non-adaptive setting (vanilla approach).
    %as already presented in \Cref{fig:T1_hom_single:subnoise}.
    The other two lines show the same setting, however, the orange line uses weight balancing for the loss terms with $\alpha=0.1$ whereas the blue line uses RBA, as already presented in \Cref{fig:T1_hom_single:subnoise}. Each setup is undertaken with ten different seeds. The solid line depicts the geometric mean over the seeds; the shaded region is the area spanned by the trajectories. The second dashed vertical line marks the selected end of training at 10k BFGS epochs, for which we report performance and associated errors. For completeness, we also display the algorithm's performance up to 50k epochs.}
    \label{fig:rba_weight_balancing_comp}
\end{figure*}

First, we investigate the effect of weight balancing, as discussed in \Cref{sec:adaptive_weights}, on the homogeneous test case from \Cref{sec:res_static}. 
As can be seen in \Cref{fig:weight_balancing}, the best result is still achieved without weight balancing, and setting the control parameter $\alpha$ for the moving average to high values leads to worse results. 
However, also note that with a small value of $\alpha$, meaning that we do not change the weights too drastically, the results are comparable to the ones without weight balancing with the advantage of a reduced variability in the reconstruction.
\begin{table}[ht]
\centering
\begin{tabular}{ccc} 
\toprule
\textbf{type} & \textbf{$L^2$ error on $\bu$ (\SI{}{\mm})} & \boldsymbol{$\epsilon_{S_a; rel}$}\\
\midrule
not adaptive & \num{9.47e-03} & \num{7.18e-02} \\ \midrule
adaptive, $\alpha=0.5$ & \num{8.05e-03} & \num{2.90e-01} \\ \midrule
adaptive, $\alpha=0.1$ & \num{7.75e-03} & \num{9.01e-02} \\ \midrule
RBA & \num{7.68e-03} & \num{4.59e-02} \\
\bottomrule
\end{tabular}
\caption{Comparison of results using adaptive weighting methods for the quasi-static case, with $S_a$ modelled as a single parameter, and incorporating noise with $LD=0.05$. 
The first column indicates what type of adaptive weighting is used and the value given to the moving average parameter $\alpha$ for the case of weight balancing. 
The second column shows $\mathcal J_{\text{OBS; test}}$, the $L^2$ (absolute) testing error for the state $\bu$ evaluated on \num{1000} randomly chosen points. 
The last column shows the relative error on $S_a$ w.r.t the ground-truth value. 
Both errors are first averaged over the the values from \num{9900} to \num{10000} iterations and then averaged over ten different seeds.}
\label{tab:weight_balancing}
\end{table}
For the sake of completeness, we also show a comparison between the vanilla approach (consisting of fixing the weights during Adam phase), the use of adaptive weights with $\alpha = 0.1$, and RBA (as defined in \Cref{sec:RBA}) on the PDE loss. 
\Cref{fig:rba_weight_balancing_comp} shows that the best estimation of $S_a$ is achieved using RBA on the PDE loss, which justifies the choice made for the test cases presented in this work.
We also compare in~\Cref{tab:weight_balancing} the $L^2$ (absolute) testing error for the state $\bu$ evaluated on \num{1000} randomly chosen points and 
the relative error on $S_a$ w.r.t the ground-truth value obtained with the vanilla approach (without adaptive weighting schemes), with weight balancing and with RBA, respectively.

\section{Comparison of results without using regularisation for the active stress network}
\label{sec:appendix_reg}
\begin{figure*}[h!]
    \centering
    \includegraphics[width=0.9\linewidth]{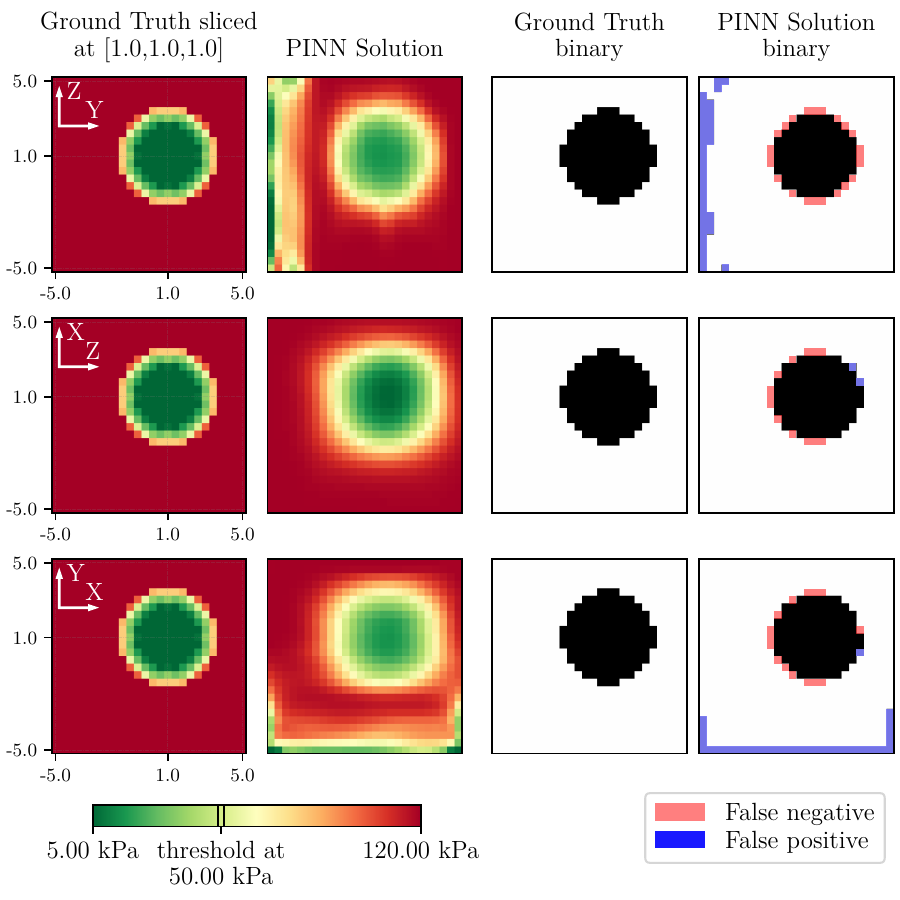}
    \caption{Field reconstruction without the modified boundary losses. The plot shows the results for the parameter reconstruction obtained with the PINN formulation using standard boundary losses and without further regularisation. The plot presents slices of the parameter field through the centre of the scar, each row representing a slice parallel to a coordinate plane. The first column shows the ground-truth parameter field, the second column the PINN reconstruction. The next two columns show binary classifications based on a threshold value of $T=\SI{50.00}{\kilo \pascal}.$ The blue and red colours in the PINN solution represent regions where the solution is wrongly classified.}
    \label{num:fig:scarregularisationnoreg}
\end{figure*}

\begin{figure*}[h!]
    \centering
    \includegraphics[width=0.9\linewidth]{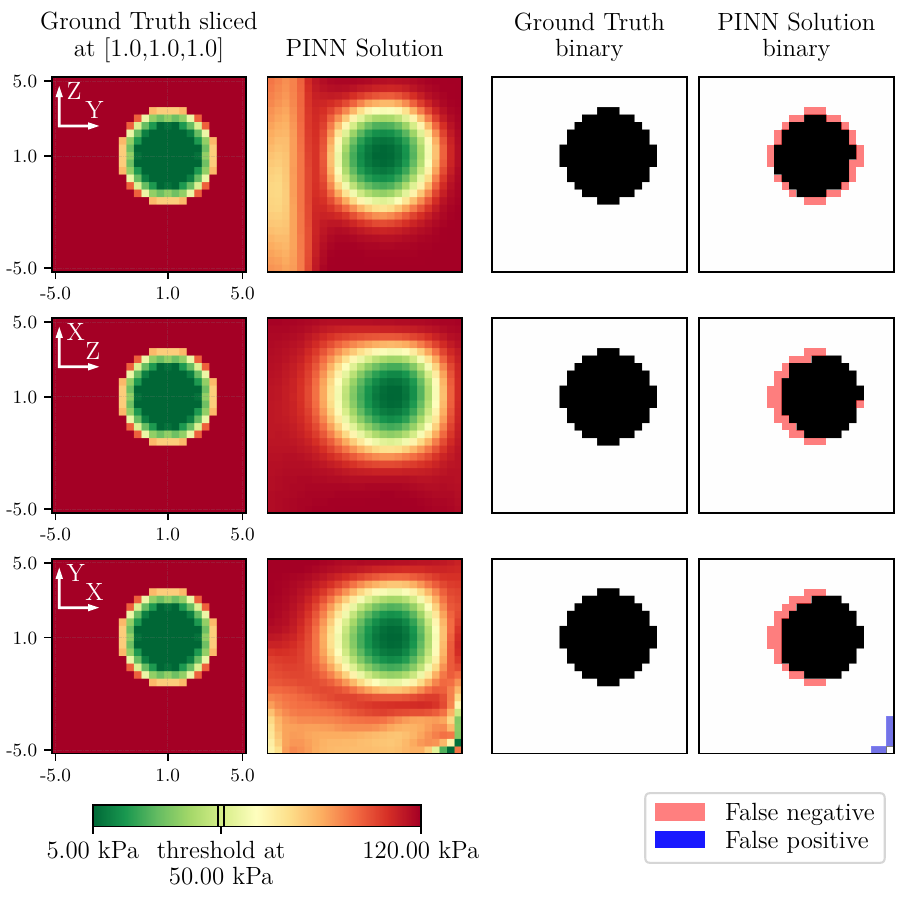}
    \caption{Field reconstruction using the modified boundary losses and the regularisation with $\lambda_{\text{REG}}=\num{1e-4}$. The plot shows the results for the parameter reconstruction obtained with the PINN formulation using the modified boundary losses and the additional regularisation for the parameter network. The plot presents slices of the parameter field through the centre of the scar, each row representing a slice parallel to a coordinate plane. The first column shows the ground-truth parameter field, the second column the PINN reconstruction. The next two columns show binary classifications based on a threshold value of $T=\SI{50.00}{\kilo \pascal}.$ The blue and red colours in the PINN solution represent regions where the solution is wrongly classified.}
    \label{num:fig:scarregularisationreg}
\end{figure*}

We illustrate the effect of the modified boundary loss and the additional regularisation in a preliminary comparison shown in Figures \ref{num:fig:scarregularisationnoreg} and \ref{num:fig:scarregularisationreg}. Both test cases use \num{6000} displacement data points and \num{1000} strain data points, corrupted with a noise at level of $LD=0.05$. 
As it can be seen in Figure \ref{num:fig:scarregularisationnoreg}, the parameter field network underestimates the solution near the Dirichlet face. This occurs because the displacement network fails to approximate the strain accurately in a low-data regime.\\
The proposed modification, shown in Figure \ref{num:fig:scarregularisationreg}, mitigates this issue to some extent. However, some artifacts remain, now more concentrated near the edges. Further improvement is possible by incorporating additional strain data, as demonstrated in subsequent test cases. Nevertheless, we retain this regularisation strategy, as it generally enhances the reconstruction quality. Additionally, this issue may primarily stem from model-related challenges on the Dirichlet face, where the displacement field exhibits singular behaviour.

% \section{Comparison of results using weight decay for active stress field}
% \label{sec:appendix_weight_decay}
% not so important - can be neglected

\section{Comparison of results using weak homogeneous Dirichlet boundary conditions}
\label{sec:appendix_BC_weak}
\begin{figure*}[h!]
    \centering
    \includegraphics[width=0.7\linewidth]{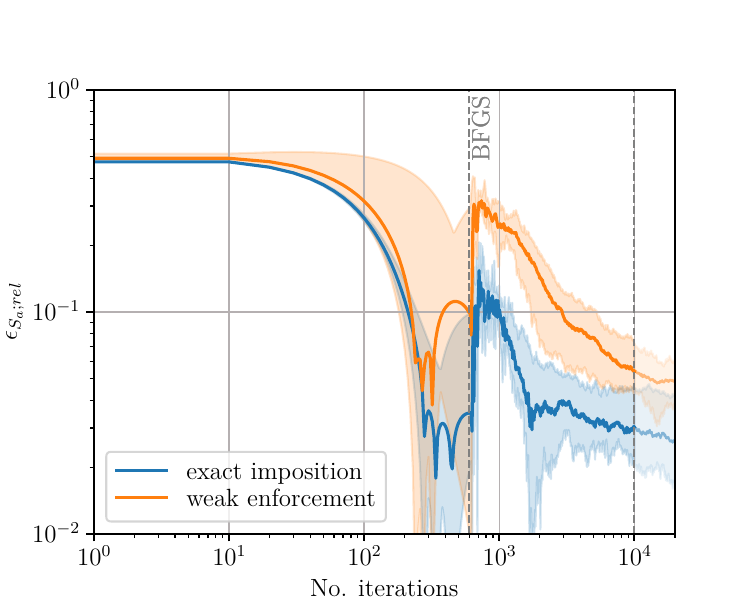}
    \caption{Comparison of results using weak enforcement of the Dirichlet boundary condition with an according loss term vs. exact imposition of the Dirichlet boundary condition in the network structure. The plot shows the relative error on the parameter $S_a$ for the quasi-static case, with $S_a$ modelled as a single parameter and without noise. The blue line uses the exact imposition whereas the orange line uses the weak enforcement of the Dirichlet boundary. The solid line depicts the geometric mean over the seeds; the shaded region is the area spanned by the trajectories. The second dashed vertical line marks the selected end of training at 10k BFGS epochs, for which we report performance and associated errors. For completeness, we also display the algorithm's behavior up to 20k epochs.}
    \label{num:fig:boundarylayercomp}
\end{figure*}

This section gives a comparison of the effect of the exact imposition of the Dirichlet boundary as in~\Cref{sec:exact_BCD} and the weak enforcement via a loss term. The hyperparameter used are those of the noiseless setting in the homogeneous test case of Section \ref{sec:res_static}. Both cases are undertaken with five different seeds. 
\Cref{tab:boundarylayercomp} displays the averaged end values for the state reconstruction and the relative error on the parameter $S_a$, whereas \Cref{num:fig:boundarylayercomp} shows a direct visual comparison of the evolution of the relative errors for both types of methods. 
Although both methods give similar results for the state and parameter reconstruction, the trajectory for the relative error descents faster for the case of the exact imposition of the Dirichlet boundary.
\begin{table}[ht]
\centering
\begin{tabular}{ccc} 
\toprule
\textbf{type} & \textbf{$L^2$ error on $\bu\, $ (\SI{}{\mm})} & \boldsymbol{$\epsilon_{S_a; rel}$}\\
\midrule
exact imposition & \num{1.01e-03} & \num{3.04e-02} \\ \midrule
weak enforcement & \num{1.14e-03} & \num{5.43e-02} \\
\bottomrule
\end{tabular}
\caption{
Comparison of results using weak enforcement of the Dirichlet boundary condition with an according loss term vs. exact imposition of the Dirichlet boundary condition in the network structure for the quasi-static case, with $S_a$ modelled as a single parameter and without noise. The first column indicates the used method. The second column shows $\mathcal J_{\text{OBS; test}}$, the $L^2$ testing error for the state $\bu$ evaluated on \num{1000} randomly chosen points. The last column shows the relative error on $S_a$ w.r.t the ground-truth value. Both errors are first averaged over the the values from \num{9900} to \num{10000} iterations and then averaged over five different seeds.
}
\label{tab:boundarylayercomp}
\end{table}

\section{Comparison of results using homogeneous Dirichlet or Robin boundary conditions}
\label{sec:appendix_BC_robin}
\begin{figure*}[h!]
    \centering
    \includegraphics[width=0.7\linewidth]{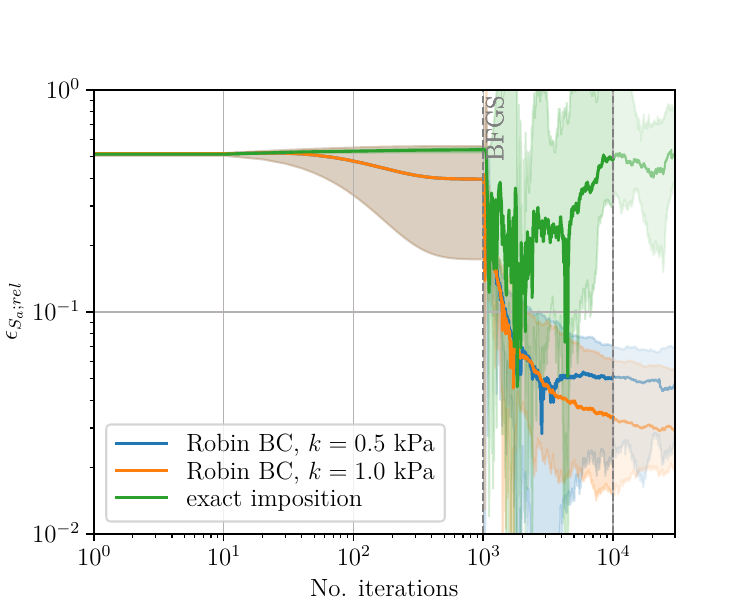}
    \caption{Comparison of results where the ground-truth solution is generated by replacing the Dirichlet face with a Robin boundary condition with a spring stiffness of $\SI{1}{\kilo\Pa}$. The plot shows the relative errors on the parameter $S_a$ for the quasi-static case, with $S_a$ modelled as a single parameter and with a noise level of $LD=\num{0.05}$. The blue line weakly enforces the Robin boundary condition, however, by using a different spring stiffness than in the ground-truth solution. The orange line uses the correct spring stiffness. The green line assumes a Dirichlet boundary condition, which was imposed exactly into the network as given in ~\Cref{sec:exact_BCD}. The solid line depicts the geometric mean over the seeds; the shaded region is the area spanned by the trajectories. The second dashed vertical line marks the selected end of training at 10k BFGS epochs, for which we report performance and associated errors. For completeness, we also display the algorithm's behavior up to 20k epochs.}
    \label{num:fig:Robincomp}
\end{figure*}

In this section we compare the results of the PINN estimation using homogeneous Dirichlet boundary conditions on the boundary $y = y_{min}$ and using as ground-truth data a FE simulation considering Robin boundary conditions with a very strong spring stiffness (namely $\SI{1}{\kilo\Pa}$), which provides a very similar mechanical response.
The goal is to investigate whether the identifiability issues at this boundary layer stem from a difficulty of PINNs in handling Dirichlet boundary conditions and the robustness of the prediction to model uncertainty on the boundary. 
~\Cref{num:fig:Robincomp} shows a comparison of the relative error on $S_a$ enforcing Robin BC with spring stiffness equal to $\SI{1}{\kilo\Pa}$ or $\SI{0.5}{\kilo\Pa}$ in the PINN training, respectively, or exactly imposing Dirichlet BC as in~\Cref{sec:exact_BCD}. 
We deduce that when Robin BC are considered for the ground-truth data and enforced in the PINN training, we obtain similar relative error on $S_a$ as in the case where homogeneous Dirichlet BC are considered for the ground-truth and exactly imposed in the PINN training, (shown in~\Cref{num:fig:boundarylayercomp}). 
In addition, these results are robust to uncertainty on the given spring stiffness.
The PINN performance is however worse when Robin BC ground-truth data are used and Dirichlet BC are exactly imposed in the PINN training.
Therefore, we conclude that the identifiability issue is instead related to the induced mechanical response, as the displacement field is primarily governed by this boundary condition.
%to show that the identifiability issues at this boundary layer are not strictly related to the Dirichlet boundary conditions per se, but to the induced mechanical response, since the displacement field is dominated by this boundary condition.

\section{Optimal threshold for classification in heterogeneous test cases}
\label{sec:appendix_threshold}
The primary goal of the reconstruction is the detection of heterogeneities (e.g. scars) and tissue classification. 
To achieve this, the PINN reconstruction undergoes a post-processing step for classification. 
We employ a simple thresholding approach, as it does not require prior topological assumptions about the solution, such as the number or shape of scars. 
However, this method introduces an additional hyperparameter — the classification threshold.

\begin{figure}[h!]
    \centering
    \includegraphics[width=0.5\textwidth]{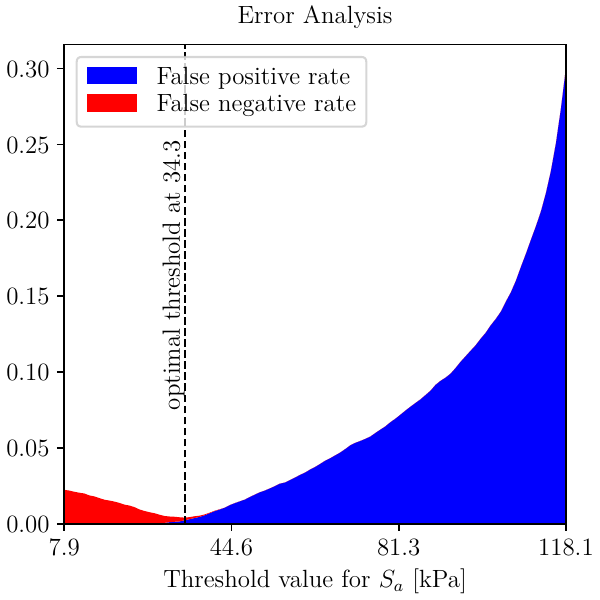}
    \caption{Single scar test case. The plot illustrates how varying the threshold value impacts the misclassification rates. False positives refer to regions where the PINN reconstruction incorrectly detects a scar, despite the tissue being healthy. False negatives refer to regions where the PINN reconstruction classifies tissue as healthy, even though a scar is actually present.}
    \label{fig:onescarbinary}
\end{figure}

\begin{figure}[h!]
    \centering
    \includegraphics[width=0.5\textwidth]{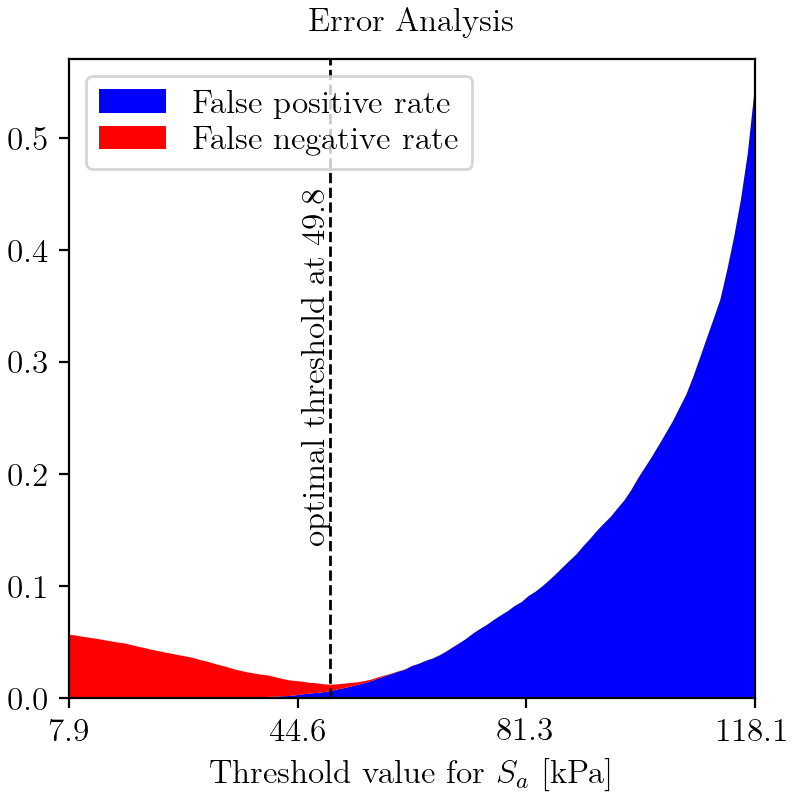}
    \caption{Two-scar test case. The plot shows how the variation of the threshold value affects the misclassification rates. False positive refers to regions where the PINN reconstruction incorrectly detects a scar. False negative refers to regions where the PINN reconstruction classifies tissue as healthy even though a scar is present.}
    \label{fig:twoscarbinary}
\end{figure}

To evaluate the classification performance, we define the False Positive (FP) and False Negative (FN) rate for a given threshold $T > 0$ as
\[
\begin{aligned}
    \text{FP} &= \{\bx \in \R^3 \mid S_a(x) = S_a^*  \, \text{ and }\, \mathrm{NN}_{S_a}(\bx; \bw_2) < T\}, \\
    \text{FN} &= \{\bx \in \R^3  \mid S_a(x) < S_a^*  \text{ and } \mathrm{NN}_{S_a}(\bx; \bw_2) \geq T\}, \\
\end{aligned}
\]
with $S_a^* = \SI{118.08}{\kilo \pascal}$.
Similarly, we define the True Positive (TP) and True Negative (TN) rate as
\[
\begin{aligned}
    \text{TP} &= \{\bx \in \R^3  \mid S_a(x) =  S_a^* \text{ and } \mathrm{NN}_{S_a}(\bx; \bw_2) \geq T\}, \\
    \text{TN} &= \{\bx \in \R^3 \mid S_a(x) <  S_a^*  \text{ and } \mathrm{NN}_{S_a}(\bx; \bw_2) < T\}. \\
\end{aligned}
\]
From these, we compute the FP rate (FPR) and FN rate (FNR) as
\[
\begin{aligned}
    \text{FPR} &= \frac{\text{FP}}{\text{FP}+\text{TN}}, \\
    \text{FNR} &= \frac{\text{FN}}{\text{FN}+\text{TP}}. \\
\end{aligned}
\]
The total misclassification rate is then given by $\text{FPR}+\text{FNR}$.\\
For the sake of completeness, Figures \ref{fig:onescarbinary} and \ref{fig:twoscarbinary} plot FPR and FNR against different threshold values $T$ for the single scar and two-scar test cases of \Cref{sec:one_scar} and \Cref{sec:two_scar}, respectively. 
The optimal threshold for both test cases is approximately $\SI{50.00}{\kilo\pascal}$. 
For the sake of simplicity, we then consider this threshold value for both test cases as a satisfying educated guess of the optimal threshold.
Note that, however, the optimal threshold depends on the relative proportions of healthy and scarred tissue in the ground truth. 
For instance, if the entire tissue were classified as healthy, the total misclassification rate could still appear relatively low.

\end{document}